\documentclass{article}

        \PassOptionsToPackage{numbers}{natbib}

\usepackage[preprint]{neurips_2026}

\usepackage[utf8]{inputenc} 
\usepackage[T1]{fontenc}    
\usepackage{hyperref}       
\usepackage{url}            
\usepackage{booktabs}       
\usepackage{amsfonts}       
\usepackage{nicefrac}       
\usepackage{microtype}      
\usepackage{xcolor}         
\usepackage{enumitem}
\usepackage{graphicx}
\usepackage{amsmath}
\usepackage{tcolorbox}
\tcbuselibrary{breakable,skins}
\usepackage{algorithm}
\usepackage{algorithmic}
\usepackage{longtable}
\usepackage{booktabs}
\usepackage{subfig}    
\usepackage{wrapfig}
\usepackage{xcolor}
\definecolor{lightblue}{HTML}{dae8fc}
\definecolor{myblue}{HTML}{6c8ebf}
\definecolor{myback}{HTML}{e9f1f2}

\newcommand{\ie}{\textit{i.e.}}

\newcommand{\model}{UniGraphLM }
\newcommand{\modelns}{UniGraphLM}
\title{A Unified Graph Language Model for Multi-Domain Multi-Task Graph Alignment Instruction Tuning}

\author{
  Haibo Chen \\
  Tsinghua University \\
  \texttt{chb24@mails.tsinghua.edu.cn} \\
  \And
  Xin Wang\thanks{Corresponding author.} \\
  Tsinghua University \\
  \texttt{xin\_wang@tsinghua.edu.cn} \\
  \And
  Jiaheng Chao \\
  Tsinghua University \\
  \texttt{chaojiaheng21@gmail.com} \\
  \And
  Ling Feng \\
  Tsinghua University \\
  \texttt{fengling@tsinghua.edu.cn} \\
  \And
  Wenwu Zhu\footnotemark[1] \\
  Tsinghua University \\
  \texttt{wwzhu@tsinghua.edu.cn} \\
}

\begin{document}

\maketitle

\begin{abstract}

Leveraging Graph Neural Networks (GNNs) as graph encoders and aligning the resulting representations with Large Language Models (LLMs) through alignment instruction tuning has become a mainstream paradigm for constructing Graph Language Models (GLMs), combining the generalization ability of LLMs with the structural modeling capacity of GNNs.  
However, existing GLMs that adopt GNNs as graph encoders largely overlook the problem of aligning GNN-encoded representations across domains and tasks with the LLM token space to obtain unified graph tokens, thereby limiting their ability to generalize across diverse graph data. 
To bridge this gap, we aim to incorporate a multi-domain, multi-task GNN encoder into GLMs and align its representations with LLMs to enable multi-domain, multi-task graph alignment instruction tuning. 
This alignment problem remains underexplored and poses two key challenges: 
1) learning GNN-encoded representations that are simultaneously generalizable across domains and tasks and well aligned with textual semantics is difficult, due to substantial variations in graph structures, feature distributions, and supervision signals, together with the lack of textual-semantic alignment guidance in task-specific GNN training; 
2) diverse graph data and task-specific instructions can exhibit different degrees of compatibility with the LLM token space during instruction tuning, leading to varying alignment difficulty and rendering a fixed alignment strategy suboptimal. 
To tackle these challenges, we propose \textbf{\modelns}, a \textbf{{U}}nified {\textbf{Graph}} \textbf{{L}}anguage \textbf{{M}}odel that incorporates a multi-domain, multi-task GNN encoder to learn generalizable graph representations aligned with textual semantics, and then adaptively aligns these representations with the LLM.
Specifically, we first develop a graph-text pair pretraining strategy with a tailored GNN encoder, trained on large-scale graph-text data spanning multiple domains and tasks to obtain generalizable representations naturally aligned with textual semantics. We further design a curriculum alignment tuning strategy that adaptively adjusts the alignment process by accounting for varying alignment difficulty across diverse graph data.
Extensive experiments demonstrate that \model consistently outperforms state-of-the-art baselines across graph datasets from different domains and tasks.
\end{abstract}

\section{Introduction}

Recent advances in Large Language Models (LLMs) have motivated the development of Graph Language Models (GLMs)~\cite{he2024g,ye2024language,chai2025graphllm}, which aim to extend the generalization and reasoning capabilities of LLMs to graph-structured data. Inspired by the success of Vision Language Models (VLMs)~\cite{liu2023visual,li2023blip,liu2024improved}, existing GLMs typically follow a VLM-style two-stage architecture: a graph encoder first maps graph-structured data into graph representations, which are then aligned with the LLM token embedding space through alignment instruction tuning to produce graph tokens for downstream tasks. In this paradigm, Graph Neural Networks (GNNs) have become the dominant choice of graph encoders due to their strong ability to capture both structural and semantic information in graphs~\cite{tang2024graphgpt,chen2024llaga}. As a result, leveraging GNNs as graph encoders and aligning their representations with LLMs has become a mainstream paradigm for constructing GLMs, combining the expressive power of GNNs on graph data with the strong generalization capabilities of LLMs.

However, existing GLMs that use GNNs as graph encoders are typically designed around task- or domain-specific graph representation learning, with limited consideration of how GNN-encoded representations from diverse domains and tasks can be consistently aligned with the LLM token space to obtain unified graph tokens. This restricts their ability to generalize across diverse graph data. Such limited generalization stems from the inherent diversity of graph data: graph structures, feature distributions, and supervision signals often vary substantially across domains and tasks, making it difficult for GNNs to learn generalizable representations in a unified manner~\cite{liu2023one,sun2025handling}. Moreover, the modality gap between GNN-encoded representations and textual semantics further complicates their alignment with LLMs~\cite{zhang2024graphtranslator,guo2023gpt4graph,wang2024llms}. In addition, such variations in graph structures, feature distributions, and task-specific instructions also lead to varying levels of alignment difficulty across graph data, making it challenging for a unified alignment strategy to adapt effectively~\cite{wang2025generalization}.

Motivated by these observations, this paper studies how to incorporate a multi-domain, multi-task GNN encoder into GLMs and align its representations with the LLM token space to enable multi-domain, multi-task graph alignment instruction tuning, producing unified graph tokens for diverse graph data. Despite its importance, this alignment problem remains underexplored and raises two key challenges:
1) \textbf{Generalizable text-aligned representation learning.} Learning GNN-encoded representations that are simultaneously generalizable across domains and tasks and well aligned with textual semantics is difficult, due to substantial variations in graph structures, feature distributions, and supervision signals, together with the lack of textual-semantic alignment guidance in task-specific GNN training.
2) \textbf{Varying alignment difficulty.} Diverse graph data and task-specific instructions can exhibit different degrees of compatibility with the LLM token space during instruction tuning, resulting in varying alignment difficulties and making a fixed alignment strategy suboptimal.

To tackle these challenges, we propose \textbf{\modelns}, a \underline{\textbf{Uni}}fied \underline{\textbf{G}}raph \underline{\textbf{L}}anguage \underline{\textbf{M}}odel that incorporates a multi-domain, multi-task GNN encoder to learn generalizable graph representations aligned with textual semantics across domains and tasks, and adaptively aligns these representations with the LLM token space during instruction tuning. 
Specifically, we first propose a graph-text pair pretraining strategy, where a tailored GNN encoder is trained on large-scale graph-text datasets spanning multiple domains and tasks, enabling it to learn generalizable representations naturally aligned with textual semantics and thereby facilitating subsequent alignment with LLMs. 
Furthermore, we design a curriculum alignment tuning strategy that adaptively adjusts the alignment process by accounting for varying alignment difficulties induced by the diversity of graph data across domains and tasks, enabling more effective alignment of GNN representations with LLMs.
Extensive experiments demonstrate that \model consistently outperforms state-of-the-art baselines across diverse graph domains and tasks. 
The contributions of this paper are summarized as follows:
\begin{itemize}[leftmargin=0.5cm]
\item To the best of our knowledge, \model is the first work to jointly incorporate a shared multi-domain, multi-task GNN encoder into GLMs and explicitly align its representations with LLMs to produce unified graph tokens for diverse graph data, paving the way for GNN-encoder-based graph language models that generalize across domains and tasks.
\item We propose a graph-text pair pretraining strategy for learning generalizable representations aligned with textual semantics, and a curriculum alignment tuning strategy for adapting the alignment process to varying difficulty across diverse graph data.
\item We conduct extensive experiments across diverse graph domains and tasks, demonstrating that \model consistently outperforms state-of-the-art GLM baselines under both multi-domain multi-task learning and cross-domain/cross-task generalization settings.
\end{itemize}

\section{Related Works}

\paragraph{LLM for Graph via Graph-to-Text.}
With the strong capabilities of Large Language Models (LLMs) in natural language understanding and reasoning, a natural approach is to convert graphs into textual descriptions, \ie, graph-to-text, and feed them into LLMs to perform graph-related tasks~\cite{tan2023walklm, tang2024grapharena, dailarge, wang2024instructgraph,yu2026graph2text}. For instance, NLGraph~\cite{wang2023can} transforms graph structures into natural language problem descriptions and evaluates whether LLMs can directly solve classical graph reasoning tasks, such as connectivity, shortest path, and maximum flow, through textual inputs. Similarly, LLM4DyG~\cite{zhang2024llm4dyg} introduces a benchmark that encodes dynamic graph structures into natural language and designs a diverse set of tasks, including temporal link prediction, path reasoning, and triadic closure, to systematically assess LLMs' ability to understand and reason over both structural and spatio-temporal information from text. However, this paradigm suffers from inherent limitations: due to the large scale and complex topology of graph data, it is difficult to faithfully represent an entire graph using plain text; moreover, encoding full graph information as textual input incurs substantial token overhead, leading to increased computational cost and limited scalability.

\paragraph{LLM for Graph via Graph-to-Token.}
In contrast to graph-to-text approaches, another line of work adopts a graph-to-token paradigm, which we refer to as Graph Language Models (GLMs)~\cite{perozzi2024let,chen2025hierarchical,guo2026realm,zhang2026toward,xu2026gnnasjudgeunleashingpowerllms,Wan_Wang_Fang_Wu_2026} in this paper. In this paradigm, a graph encoder transforms graph structures into compact continuous representations, which are aligned with the LLM token embedding space for downstream reasoning~\cite{wang2024graph2token,wangunigte,li2026onesizefitsalladaptivesubgraphdenoising}. Compared with verbose natural language descriptions, such graph tokens enable more efficient and scalable processing of graph data. 
For example, GraphGPT~\cite{tang2024graphgpt} aligns graph structural knowledge with LLMs through text-graph grounding and dual-stage instruction tuning, enabling generative graph reasoning. Similarly, LLaGA~\cite{chen2024llaga} adapts graph data into a structure-aware sequential format by reorganizing nodes and projecting graph representations into the LLM token embedding space, enabling unified sequence modeling over graph inputs. 
Owing to their strong capability in modeling graph-structured data, Graph Neural Networks (GNNs) have become the dominant choice of graph encoders, transforming graphs into expressive continuous representations~\cite{tang2024graphgpt,chen2024llaga, zhang2024graphtranslator,wang2024llms}. However, existing GNN-based GLMs are typically trained on a single domain or task, and still face significant challenges in aligning multi-domain, multi-task GNN representations with the LLM token space to obtain unified graph tokens for diverse graph data.

\section{Problem Formulation}
In this section, we first define multi-domain, multi-task graph data and then describe the graph alignment instruction tuning process. Finally, we formulate the problem of multi-domain, multi-task graph alignment instruction tuning. 

\subsection{Multi-domain, Multi-task Graph Data}

We consider a collection of graph datasets $\mathcal{D}$, where each dataset $D \in \mathcal{D}$ is treated as a distinct domain and is associated with a task type $t_D \in \mathcal{T}$, with $\mathcal{T}$ denoting the task-type space.
The task type determines the required granularity of the graph representation (node-, edge-, or graph-level).
Formally, dataset $D$ is represented as a set of graph-task instances $\{(G_i, y_i)\}_{i=1}^{N_D}$, where $G_i$ is a graph instance drawn from $D$, $y_i \in \mathcal{Y}_{t_D}$ is its task-specific label, and $N_D$ is the number of instances in $D$.
Each graph instance is defined as $G_i = (\mathcal{V}_i, \mathcal{E}_i, \mathbf{X}_i, \mathbf{E}_i)$, where $\mathcal{V}_i$ and $\mathcal{E}_i$ denote its node and edge sets, and $\mathbf{X}_i$, $\mathbf{E}_i$ denote the corresponding node and edge features.

\subsection{Graph Alignment Instruction Tuning}

For a single dataset $D$ with task type $t_D$, graph alignment instruction tuning converts each graph-task instance $(G_i, y_i)$ into a graph-conditioned instruction. 
Let $\hat{\mathbf{X}}_i$ denote the task-granularity graph representation obtained by applying the graph encoder parameterized by $\phi$ to $G_i$. 
A projector layer then maps $\hat{\mathbf{X}}_i$ into a sequence of graph tokens $\mathbf{z}_i = \mathrm{Project}_{\theta}(\hat{\mathbf{X}}_i)$. 
The graph-conditioned instruction is then constructed as $I_i = [\mathbf{q}_i; \mathbf{z}_i]$, where $\mathbf{q}_i$ denotes the natural language instruction describing the task and $\mathbf{z}_i$ denotes the graph token sequence. 
The target output $\mathbf{t}_i$ is derived from the task label $y_i$. The instruction tuning objective follows the standard next-token prediction paradigm, where the LLM generates $\mathbf{t}_i$ autoregressively conditioned on the instruction $\mathbf{q}_i$ and graph tokens $\mathbf{z}_i$:
\begingroup
\small
\begin{align}
\mathcal{L}_{D} = - \sum_{i=1}^{N_D} 
\log P(\mathbf{t}_i \mid \mathbf{q}_i, \mathbf{z}_i; \phi, \theta, \psi),
\label{eq:instruction-tuning}
\end{align}
\endgroup
where $\phi$ denotes the parameters of the graph encoder, $\theta$ denotes the parameters of the projector layer, and $\psi$ denotes the parameters of the LLM.

\subsection{Multi-domain, Multi-task Graph Alignment Instruction Tuning}

This work studies how to incorporate a multi-domain, multi-task GNN encoder into GLMs and jointly align its representations from multiple datasets with LLMs to produce unified graph tokens for diverse graph data. 
Since different graph tasks require different representation granularities, a unified encoder should be able to produce node-level, edge-level, and graph-level representations, denoted as $\mathbf{X}_i^{\text{node}}(v)$, $\mathbf{X}_i^{\text{edge}}(u,v)$, and $\mathbf{X}_i^{\text{graph}}$, respectively. For each dataset $D \in \mathcal{D}$ with task type $t_D$, let $\ast \in \{\text{node}, \text{edge}, \text{graph}\}$ denote the required representation granularity. A multi-scale GNN encoder $\mathrm{GNN}^{\text{multi}}_{\phi}(\cdot)$ extracts the task-specific representation $\mathbf{X}_i^{\ast} = \mathrm{GNN}^{\text{multi}}_{\phi}(G_i, \ast)$, which is then projected into the LLM token space to obtain graph tokens $\mathbf{z}_i^{\ast} = \mathrm{Project}_{\theta}(\mathbf{X}_i^{\ast})$.

The goal of multi-domain, multi-task graph representation alignment is to learn a shared GNN encoder and projector that can produce graph tokens consistently aligned with LLMs across all datasets in $\mathcal{D}$. Accordingly, the overall alignment objective aggregates the instruction tuning losses over multiple datasets:
\begingroup
\small
\begin{align}
\mathcal{L}_{\text{align}} = \sum_{D \in \mathcal{D}} \mathcal{L}_{D}
= - \sum_{D \in \mathcal{D}} \sum_{i=1}^{N_D} 
\log P(\mathbf{t}_i \mid \mathbf{q}_i, \mathbf{z}_i^{\ast}; \phi, \theta, \psi),
\label{eq:multi-domain-alignment}
\end{align}
\endgroup
where $\mathcal{L}_{D}$ is the alignment loss on dataset $D$. This requires the GNN encoder to learn representations that are both generalizable across domains and tasks and amenable to alignment with LLMs, while the projector should adapt the alignment process to graph data with varying alignment difficulties.

\section{Method}
\begin{figure}[!tbp]
\centering
\includegraphics[width=0.99\textwidth]{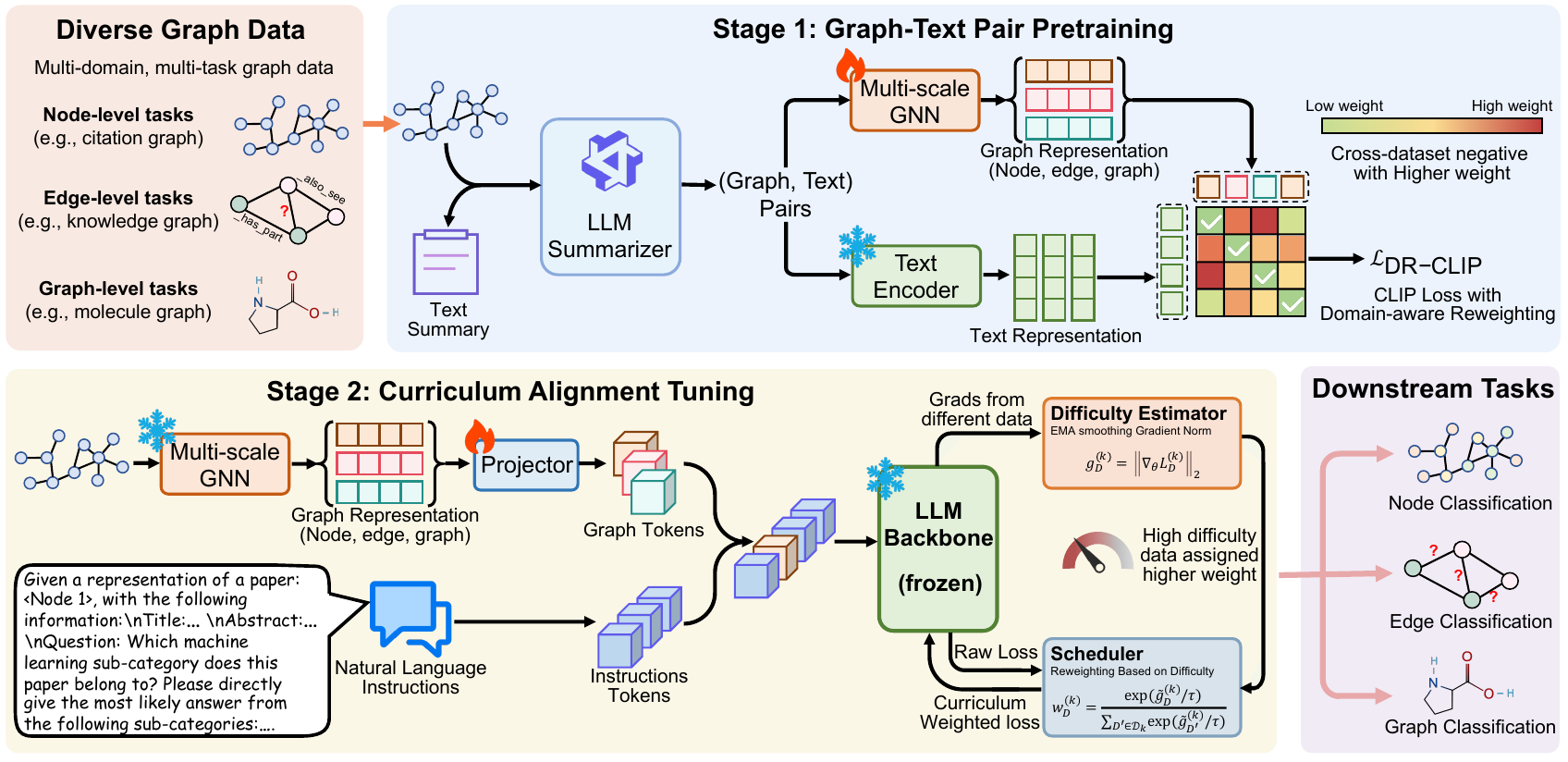}
\caption{Overall framework of \modelns. \textit{Stage 1: Graph-Text Pair Pretraining.} We construct large-scale graph-text pairs across multiple domains and tasks, encode each graph using a multi-scale GNN encoder to produce its task-required node-, edge-, or graph-level representation in a shared space, and train the encoder with a domain-aware reweighted contrastive objective that explicitly accounts for both inter-domain and intra-domain semantic differences. 
\textit{Stage 2: Curriculum Alignment Tuning.} During instruction tuning, we estimate domain-level alignment difficulty online from per-domain gradient statistics and adaptively reweight the training objective to focus more on harder domains, leading to more balanced and effective alignment between GNN representations and the LLM.
}
\label{fig:framework} 
\end{figure}

In this section, we present \model in detail. It consists of two key components: graph-text pair pretraining and curriculum alignment tuning. The overall framework is illustrated in Figure~\ref{fig:framework}.

\subsection{Graph-Text Pair Pretraining} 
To learn generalizable, text-aligned representations and facilitate subsequent alignment with LLMs, we propose a graph-text pair pretraining strategy. Specifically, we construct large-scale graph-text pairs across diverse domains and tasks, design a multi-scale GNN encoder to capture representations at different granularities, and pretrain it using a contrastive objective with reweighting to capture multi-domain semantic differences and enhance generalization. 

\paragraph{Graph-Text Pair Construction.}
Initially, we construct a large-scale graph-text pair dataset by collecting graph data from multiple domains and tasks, and pairing each graph instance with a corresponding textual description. Formally, the graph-text pair dataset is defined as $\mathcal{D}_{\text{gt}} = \{(G_i, T_i)\}_{i=1}^{N}$, where $G_i$ is a graph instance and $T_i$ is its textual description. Notably, this construction does not require task labels $y_i$, making it well-suited for real-world pretraining scenarios where large-scale graph data often lacks annotations.

Inspired by GraphCLIP~\cite{zhu2025graphclip}, we use an LLM (Qwen3-8B~\cite{yang2025qwen3}) as a preprocessing tool to generate textual descriptions for graph instances. Given a graph instance $G_i = (\mathcal{V}_i, \mathcal{E}_i, \mathbf{X}_i, \mathbf{E}_i)$, the LLM summarizes its raw textual features $(\mathbf{X}_i, \mathbf{E}_i)$ while incorporating structural information $(\mathcal{V}_i, \mathcal{E}_i)$, including node and edge attributes, to produce a coherent natural language description. We design dataset-specific prompts to generate descriptions tailored to the characteristics of different datasets; the full set of summarization prompts is provided in Appendix~\ref{app:summarization_prompts}. The resulting descriptions capture both structural and semantic information of each graph instance, providing rich supervision for subsequent graph-text pair pretraining.

\paragraph{Multi-scale Graph Representation.}
Since different tasks require graph representations at different scales, we design a multi-scale GNN encoder that produces representations at varying levels of granularity within a shared embedding space. First, we employ a GNN to extract node-level representations, and then derive a graph-level representation via a pooling operation. Given a graph $G_i = (\mathcal{V}_i, \mathcal{E}_i, \mathbf{X}_i, \mathbf{E}_i)$, the representations are computed as:
\begingroup
\small
\begin{align}
\mathbf{H}^{\text{node}}_i = \mathrm{GNN}(G_i), \quad
\mathbf{H}^{\text{graph}}_i = \mathrm{Pooling}(\mathbf{H}^{\text{node}}_i), 
\label{eq:node_graph_representation}
\end{align}
\endgroup
where $\mathbf{H}^{\text{node}}_i \in \mathbb{R}^{|\mathcal{V}_i| \times d}$ and $\mathbf{H}^{\text{graph}}_i \in \mathbb{R}^{d}$ denote the node-level and graph-level representations, respectively, and $d$ is the representation dimension. 

To enable a single GNN to produce representations at different levels of granularity within the same space, we further define task-specific aggregation functions that combine node- and graph-level representations into appropriate task-scale representations.

\begin{itemize}[leftmargin=0.5cm]
\item \textit{Node-level Tasks.} 
For node-level tasks, the representation of a target node $v$ combines its local node-level embedding with the global graph context: 
\begingroup
\small
\begin{align}
\mathbf{X}^{\text{node}}_i(v) = \mathrm{MLP}\left([\mathbf{H}^{\text{node}}_i(v) \, \| \, \mathbf{H}^{\text{graph}}_i]\right).
\label{eq:node-level}
\end{align}
\endgroup

\item \textit{Edge-level Tasks.} 
For edge-level tasks, we derive an edge representation by aggregating its endpoint node embeddings $(u, v)$ and incorporating global graph context:
\begingroup
\small
\begin{align}
\mathbf{X}^{\text{edge}}_i(u, v) = \mathrm{MLP}\left(\left[\frac{\mathbf{H}^{\text{node}}_i(u) + \mathbf{H}^{\text{node}}_i(v)}{2} \, \| \, \mathbf{H}^{\text{graph}}_i\right]\right), \quad (u, v) \in \mathcal{E}_i.
\label{eq:edge-level}
\end{align}
\endgroup

\item \textit{Graph-level Tasks.} 
For graph-level tasks, we apply the same aggregation module to the graph-level representation to ensure consistency in both dimensionality and embedding space: 
\begingroup
\small
\begin{align}
\mathbf{X}^{\text{graph}}_i = \mathrm{MLP}\left([\mathbf{H}^{\text{graph}}_i \, \| \, \mathbf{H}^{\text{graph}}_i]\right).
\label{eq:graph-level}
\end{align}
\endgroup
\end{itemize}

We denote the unified multi-scale GNN encoder as $\mathrm{GNN}^{\text{multi}}_{\phi}(\cdot)$, which produces representations $\mathbf{X}_i^{\ast}$ at different levels of granularity with $\ast \in \{\text{node}, \text{edge}, \text{graph}\}$.
This unified design ensures that representations at different granularities share a consistent dimensionality and embedding space, facilitating their alignment with LLMs across diverse tasks.

\paragraph{Domain-aware Reweighting.}
A natural way to train the GNN encoder on graph-text pairs is to adopt a contrastive learning objective similar to CLIP~\cite{radford2021learning}, which pulls matched graph-text pairs closer and pushes mismatched ones apart in a shared embedding space. However, since graph-text pairs are drawn from multiple domains, each corresponding to a distinct dataset, treating all negative samples equally is suboptimal: negatives from the same domain as the anchor are typically semantically closer than those from other domains, and uniform weighting may therefore over-penalize intra-domain negatives while under-utilizing inter-domain negatives. To address this issue, we introduce a domain-aware reweighting strategy that assigns larger weights to negatives from more distant domains and smaller weights to intra-domain or nearby-domain negatives, thereby encouraging stronger cross-domain separation while preserving fine-grained intra-domain semantics.

Initially, we employ GraphSAGE~\cite{hamilton2017inductive} as the multi-scale GNN encoder to extract graph representations $\mathbf{X}_i^{\ast}$ (Eqs.~\ref{eq:node-level}--\ref{eq:graph-level}), and use Sentence-BERT~\cite{reimers2019sentence} as the text encoder to obtain text representations $\mathbf{T}_i$ from the textual descriptions $T_i$:
\begingroup
\small
\begin{align}
\mathbf{T}_i = \mathrm{Pooling}\big(\mathrm{Sentence\mbox{-}BERT}(T_i)\big).
\label{eq:text_representation}
\end{align}
\endgroup
Given graph-text representation pairs $\{(\mathbf{X}_i^{\ast}, \mathbf{T}_i)\}_{i=1}^N$ derived from paired inputs $\{(G_i, T_i, d_i)\}_{i=1}^N$, where $d_i \in \mathcal{D}$ denotes the domain (dataset) identifier of the $i$-th pair, we compute their cosine similarities and define the bidirectional similarity matrices as:
\begingroup
\small
\begin{align}
\mathbf{S}^{g\rightarrow t}_{ij} &= \frac{{\mathbf{X}_i^{\ast}}^\top \mathbf{T}_j}{\|{\mathbf{X}_i^{\ast}}\|_2\|\mathbf{T}_j\|_2}, \quad
\mathbf{S}^{t\rightarrow g}_{ij} = \frac{\mathbf{T}_i^\top {\mathbf{X}_j^{\ast}}}{\|\mathbf{T}_i\|_2\|{\mathbf{X}_j^{\ast}}\|_2}.
\label{eq:similarity_matrices}
\end{align}
\endgroup
To construct domain-aware weights, we compute domain-level centers using the initial graph features $\mathbf{X}_i$ of $G_i = (\mathcal{V}_i, \mathcal{E}_i, \mathbf{X}_i, \mathbf{E}_i)$ and the text representations $\mathbf{T}_i$ of the corresponding textual descriptions. These centers serve as proxies for domain similarity. For each domain $a \in \mathcal{D}$, we sample up to 1000 instances to compute graph and text centers, denoted as $\mathbf{c}^{g}_{a}$ and $\mathbf{c}^{t}_{a}$, respectively. Then, for any pair of domains $a,b \in \mathcal{D}$, we define the normalized inter-domain distances: 
\begingroup
\small
\begin{align}
[M_g]_{ab} &= \frac{1-\cos(\mathbf{c}^{g}_{a},\mathbf{c}^{g}_{b})}{\max_{u,v}(1-\cos(\mathbf{c}^{g}_{u},\mathbf{c}^{g}_{v}))}, \quad
[M_t]_{ab} = \frac{1-\cos(\mathbf{c}^{t}_{a},\mathbf{c}^{t}_{b})}{\max_{u,v}(1-\cos(\mathbf{c}^{t}_{u},\mathbf{c}^{t}_{v}))}.
\label{eq:domain_distance}
\end{align}
\endgroup
We then construct domain-aware weights as:
\begingroup
\small
\begin{align}
W^g_{ab} = 1 + [M_g]_{ab}, \quad
W^t_{ab} = 1 + [M_t]_{ab}.
\label{eq:domain_weights}
\end{align}
\endgroup
Building upon the standard contrastive formulation, we incorporate these weights into the loss. For each anchor $i$, $(i,i)$ forms the positive pair, while $j \neq i$ correspond to negative samples. The weighted contrastive losses are defined as:
\begingroup
\small
\begin{align}
\mathcal{L}_{g\rightarrow t} &=
-\frac{1}{N}\sum_{i=1}^{N}
\log \frac{\exp(\mathbf{S}^{g\rightarrow t}_{ii})}
{\exp(\mathbf{S}^{g\rightarrow t}_{ii})+\sum_{j\neq i} W^{g}_{d_i,d_j}\exp(\mathbf{S}^{g\rightarrow t}_{ij})},\\
\mathcal{L}_{t\rightarrow g} &=
-\frac{1}{N}\sum_{i=1}^{N}
\log \frac{\exp(\mathbf{S}^{t\rightarrow g}_{ii})}
{\exp(\mathbf{S}^{t\rightarrow g}_{ii})+\sum_{j\neq i} W^{t}_{d_i,d_j}\exp(\mathbf{S}^{t\rightarrow g}_{ij})},\\
\mathcal{L}_{\text{DR-CLIP}} &= \frac{1}{2}\left(\mathcal{L}_{g\rightarrow t}+\mathcal{L}_{t\rightarrow g}\right).
\label{eq:dr_clip_loss}
\end{align}
\endgroup
This design assigns larger weights to negatives drawn from more distant domains, encouraging stronger inter-domain separation while preserving fine-grained intra-domain alignment. The overall pretraining objective is $\mathcal{L}_{\text{DR-CLIP}}$, which provides generalizable, text-aligned graph representations for the subsequent instruction tuning stage.

\subsection{Curriculum Alignment Tuning}
To adaptively adjust the alignment process to varying difficulties induced by the diversity of graph data across domains and tasks, we propose a curriculum alignment tuning strategy. It estimates domain-level alignment difficulty online during instruction tuning and accordingly reweights the training objective, encouraging the model to focus more on challenging domains while maintaining exposure to easier ones for more balanced and effective alignment.

\paragraph{Alignment Difficulty Estimation.}
Due to memory and compute constraints, it is infeasible to re-estimate the alignment difficulty of every domain over the entire dataset at each optimization step. Instead, we adopt an online estimation strategy with Exponential Moving Average (EMA)~\cite{haynes2012exponential, morales2024exponential} smoothing that dynamically combines statistics from the current mini-batch with historical estimates.

At optimization step $k$, let the mini-batch be $\mathcal{B}_k$. We group samples by their originating datasets, each corresponding to a domain, yielding the set of active domains $\mathcal{D}_k$. We compute the per-instance loss $L_i$ according to Eq.~\eqref{eq:instruction-tuning}, and aggregate it to obtain the dataset-level loss for each dataset $D \in \mathcal{D}_k$:
\begingroup
\small
\begin{align}
{ L_D^{(k)} = \frac{1}{|\mathcal{B}_k^D|}\sum_{i\in\mathcal{B}_k^D} L_i,}
\label{eq:dataset_loss}
\end{align}
\endgroup
where $\mathcal{B}_k^D \subset \mathcal{B}_k$ denotes the subset of samples from dataset $D$. Then, we quantify the alignment difficulty of each domain (i.e., dataset $D$) using the gradient norm:
\begingroup
\small
\begin{align}
g_D^{(k)} = \left\|\nabla_{\theta} L_D^{(k)}\right\|_2,
\label{eq:alignment_difficulty}
\end{align}
\endgroup
where $\theta$ denotes the projector parameters. Intuitively, a larger gradient norm on the projector indicates that the current model is less aligned with the domain and thus requires a larger update, which we use as a proxy for alignment difficulty~\cite{wang2021survey}. To obtain stable estimates, we employ a two-stage smoothing scheme. Because early-stage gradients are often highly noisy, directly applying EMA may propagate this noise. We therefore use a running mean as a warmup estimator before switching to EMA. Let $T$ denote the total number of training steps and $\rho \in [0,1]$ the warmup ratio, with $T_w = \lfloor \rho T \rfloor$. The smoothed difficulty score is defined as:
\begingroup
\small
\begin{align}
\tilde g_D^{(k)} =
\begin{cases}
\mu_D^{(k)} = \frac{\sum_{u=1}^k g_D^{(u)}\mathbb{I}[D \in \mathcal{D}_u]}{\sum_{u=1}^k \mathbb{I}[D \in \mathcal{D}_u]}, & k < T_w,\\
\beta\,\tilde g_D^{(k-1)} + (1-\beta)\,g_D^{(k)}, & k \ge T_w,
\end{cases}
\label{eq:smoothed_difficulty}
\end{align}
\endgroup
where $\beta$ is the EMA momentum. For a newly observed dataset, we initialize $\tilde g_D$ with $\mu_D^{(k)}$ at its first occurrence. Importantly, the running-mean and EMA updates are performed only for active domains $D \in \mathcal{D}_k$ that appear in the current mini-batch. For inactive domains $D \notin \mathcal{D}_k$, we keep their previous estimates unchanged, i.e., $\tilde g_D^{(k)}=\tilde g_D^{(k-1)}$, rather than treating their missing gradients as zero. This procedure yields robust estimates of domain-level alignment difficulty.

\paragraph{Alignment Curriculum Schedule.}
Based on the estimated difficulty, we adaptively reweight the active domains in the current mini-batch via a temperature-scaled softmax:
\begingroup
\small
\begin{align}
w_D^{(k)} = \frac{\exp\big(\tilde g_D^{(k)}/\tau\big)}{\sum_{D'\in\mathcal{D}_k}\exp\big(\tilde g_{D'}^{(k)}/\tau\big)},
\label{eq:curriculum_weights}
\end{align}
\endgroup
where $\tau$ is a temperature hyperparameter. This formulation prioritizes domains with higher estimated difficulty, allocating more learning capacity to challenging data while still maintaining exposure to easier ones. Compared to uniform training, such adaptive reweighting mitigates the dominance of easy domains and reduces the risk of underfitting harder ones. Finally, the training objective at step $k$:
\begingroup
\small
\begin{align}
\mathcal{L}_k = \sum_{D\in\mathcal{D}_k} w_D^{(k)}\,L_D^{(k)}.
\label{eq:curriculum_loss}
\end{align}
\endgroup
This curriculum-driven objective dynamically adjusts the optimization weights assigned to different domains over the course of training, resulting in more balanced and effective alignment across diverse domains and tasks. During this stage, we freeze both the pretrained GNN encoder and the LLM, and optimize only the projector parameters. This preserves the generalizable graph representations learned during graph-text pair pretraining, allowing the curriculum strategy to focus on adapting the alignment module to the LLM. The full algorithm is provided in Appendix~\ref{app:algorithm}.

\section{Experiments}

\subsection{Experimental Setup}
\paragraph{Datasets.} We conduct experiments on diverse datasets spanning multiple domains and task types. 
For node-level tasks, we consider citation networks, including \textit{Cora}, \textit{PubMed}, and \textit{Arxiv}, as well as the web hyperlink network \textit{Wiki-CS}. 
For edge-level tasks, we adopt two knowledge graph datasets, \textit{WN18RR} and \textit{FB15K237(10-way)}. Since \textit{FB15K237} contains 237 relation labels, including all candidates in each instruction would make the input overly long; we therefore use 10-way candidate sampling, consisting of the correct label and 9 randomly sampled negative labels. 
For graph-level tasks, we use two molecular datasets, \textit{ChemHIV} and \textit{ChemPCBA}. 
Details of dataset statistics and preprocessing are provided in Appendix~\ref{app:datasets}.

\paragraph{Baselines.} 
We compare \model with pure LLM baselines and several state-of-the-art graph language models, including GraphGPT~\cite{tang2024graphgpt}, LLaGA~\cite{chen2024llaga}, GOFA~\cite{konggofa}, and TEA-GLM~\cite{wang2024llms}. To ensure a fair comparison, we use LLM backbones of comparable scale across all methods. Detailed descriptions of the baselines and their configurations are provided in Appendix~\ref{app:baselines}.

\subsection{Main Results}
To evaluate whether large-scale pretraining and alignment enable our model to generalize effectively across domains and tasks, we consider three evaluation settings: multi-domain multi-task learning, cross-domain generalization, and cross-task generalization. 

\begin{table}[!htbp]
\centering
\caption{Performance comparison of different methods on node classification, edge classification, and graph classification tasks under multi-domain and multi-task learning. The highest result is \textbf{bold}, and the second highest result is \underline{underlined}.}

\resizebox{\textwidth}{!}{
\begin{tabular}{lccccccccc}
\toprule

& \multicolumn{4}{c}{\textbf{Node Classification(ACC\%)}} 
& \multicolumn{2}{c}{\textbf{Edge Classification(ACC\%)}} 
& \multicolumn{2}{c}{\textbf{Graph Classification(AUC\%)}} \\

\cmidrule(lr){2-5} \cmidrule(lr){6-7} \cmidrule(lr){8-9} 

Method
& Cora 
& PubMed 
& Wiki-CS 
& Arxiv 
& WN18RR 
& FB15K237 
& ChemPCBA 
& ChemHIV
& \textit{\textbf{Avg.}}
\\

\midrule

Vicuna-7B
& 56.67
& 78.92
& 62.34
& 43.75
& 34.84
& 56.01
& 46.47
& 30.36
& 51.17
\\

LLaMA-2-7B
& 29.45
& 25.36
& 22.22
& 6.33
& 29.16
& 20.60
& 51.12
& 50.09
& 29.29
\\

GraphGPT
& 57.59
& 73.45
& 49.20
& 67.63
& N.S.
& N.S.
& N.S.
& N.S.
& 61.97
\\

LLaGA
& \underline{74.71} 
& 68.54 
& \underline{69.16} 
& \underline{73.83}   
& N.S.
& N.S.
& N.S.
& N.S.
& 71.56
\\

GOFA
& 63.88
& 64.72
& 66.77
& 66.11
& 21.95
& 54.41
& 60.61
& 49.78
& 56.03
\\

TEA-GLM
& 62.38 
& \underline{85.13} 
& 55.19 
& 66.81   
& \underline{92.85} 
& \underline{91.85}  
& \underline{80.40} 
& \underline{67.13} 
& \underline{75.22}
\\

\midrule

Ours
& \textbf{76.98}
& \textbf{90.49}
& \textbf{76.65}
& \textbf{74.22}
& \textbf{93.52}
& \textbf{91.96}
& \textbf{81.69}
& \textbf{70.09}
& \textbf{81.95}
\\

\bottomrule
\end{tabular}
}
\label{tab:main_results}
\end{table}
\paragraph{Multi-domain and multi-task learning.} We first assess the effectiveness of jointly learning from multiple domains and task types. Specifically, we pretrain the graph encoder on the full collection of datasets spanning diverse domains and tasks in a label-free manner, perform instruction tuning of the projector layer using the combined training sets from all datasets, and directly evaluate the resulting model on each test split without dataset-specific fine-tuning.after joint instruction tuning. 

As shown in Table~\ref{tab:main_results}, pure LLM baselines perform substantially worse than GLM methods on average (e.g., Vicuna-7B: 51.17\% and LLaMA-2-7B: 29.29\%), highlighting the importance of explicitly incorporating graph structural information via a dedicated graph encoder. Among GLMs, \model achieves the best average performance (81.95\%), outperforming the strongest baseline, TEA-GLM (75.22\%), by 6.73 percentage points. It delivers consistent gains across node-level, edge-level, and graph-level classification tasks and attains the best results on all datasets. These results validate the effectiveness of the proposed method in jointly learning from multiple domains and task types and generalizing effectively across domains and tasks.

\begin{table}[!htbp]
\centering
\caption{Performance comparison under cross-domain and cross-task settings. The best results are in \textbf{bold}, and the second best are \underline{underlined}.}

\resizebox{0.6\textwidth}{!}{
\begin{tabular}{lccccc}
\toprule

& \multicolumn{3}{c}{\textbf{Cross-domain(ACC\%)}} 
& \multicolumn{2}{c}{\textbf{Cross-task(ACC\%)}} \\

\cmidrule(lr){2-4} \cmidrule(lr){5-6} 

Method
& Cora 
& PubMed 
& Wiki-CS 
& WN18RR 
& FB15K237 
\\

\midrule

GraphGPT
& 31.53
& 38.99
& 37.78
& -
& - 
\\

LLaGA
& 13.01
& 25.74
& 5.64
& -
& - 
\\

GOFA
& 16.30
& \underline{51.14}
& \underline{40.11}
& 18.79
& \underline{16.93}
\\

TEA-GLM
& \underline{43.33}
& 34.33
& 36.69
& \underline{22.65}
& 11.72
\\

\midrule

Ours
& \textbf{55.17}
& \textbf{77.31}
& \textbf{63.93}
& \textbf{24.12}
& \textbf{55.27}
\\

\bottomrule
\end{tabular}
}
\label{tab:cross}
\end{table}
\paragraph{Cross-domain and cross-task generalization.}
For cross-domain generalization, we evaluate whether the pretrained GNN encoder can support generalization across domains when instruction tuning is performed on only a single dataset. Specifically, we use the GNN encoder obtained from multi-dataset graph-text pair pretraining, perform instruction tuning only on Arxiv, a node-level dataset, and evaluate the resulting model on other node-level datasets from different domains, including Cora, PubMed, and Wiki-CS. For cross-task generalization, we further assess whether the same pretrained GNN encoder can transfer across task types under the same single-dataset instruction tuning setting. Specifically, we again perform instruction tuning only on Arxiv, and evaluate the model on edge-level datasets, including WN18RR and FB15K237.
 
As shown in Table~\ref{tab:cross}, \model substantially improves cross-domain generalization, achieving 55.17\%, 77.31\%, and 63.93\% on Cora, PubMed, and Wiki-CS, respectively, outperforming all baselines. It also demonstrates strong cross-task transfer, especially on FB15K237 (55.27\%). These results suggest that large-scale graph-text pair pretraining equips the GNN encoder with generalizable representations, enabling effective cross-domain and cross-task generalization.

\subsection{Ablation Study}
\begin{wrapfigure}{r}{0.5\textwidth} 
    \vspace*{-0.4cm}
    \centering
  \includegraphics[width=0.5\textwidth]{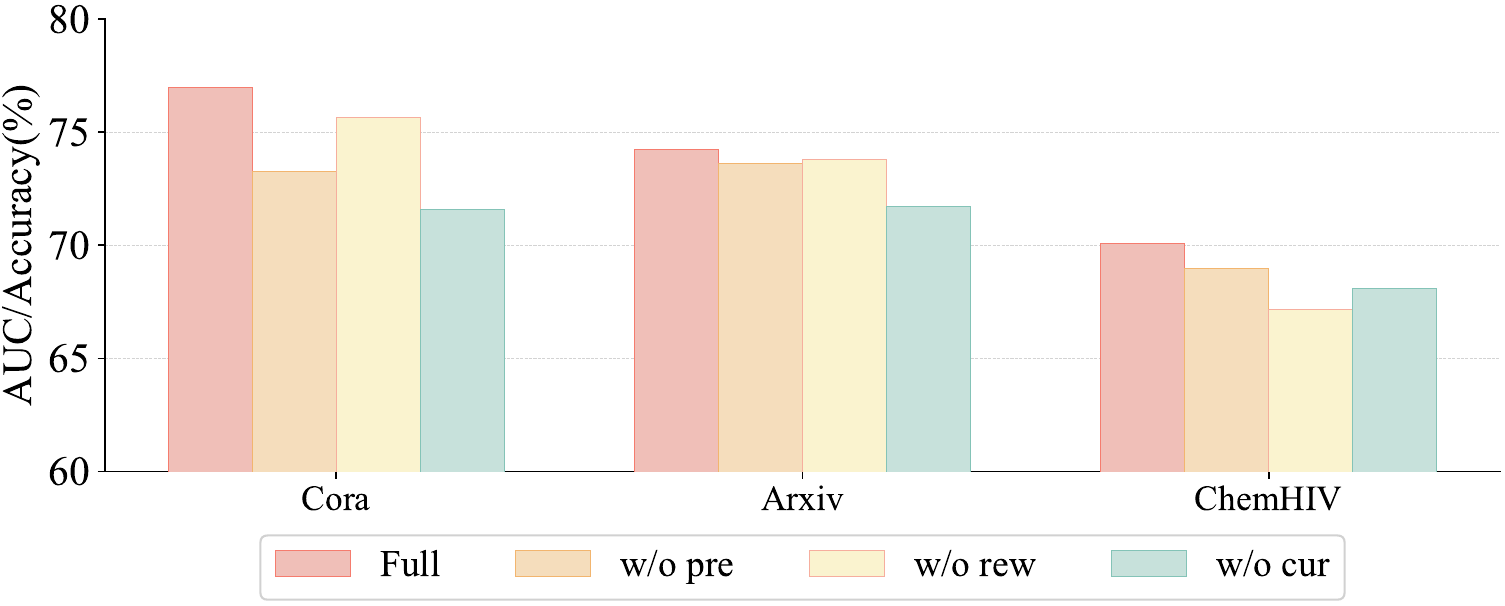}
  \caption{Performance comparison between the full model and different ablated versions.}
  \label{fig:ablation}
  \vspace*{-0.5cm}
\end{wrapfigure}
To verify the effectiveness of the proposed components, we conduct ablation studies to compare the full model with ablated versions: 1) \textbf{w/o pre}: we remove the graph-text pair pretraining, where the GNN encoder is trained along with the projector layer during instruction tuning; 2) \textbf{w/o rew}: we remove the domain-aware reweighting during graph-text pair pretraining, treating all negatives equally; 3) \textbf{w/o cur}: we remove the curriculum strategy during instruction tuning. 

The results are shown in Figure~\ref{fig:ablation}. We make the following observations. 
i) \textbf{w/o pre} exhibits a performance drop across datasets, indicating that large-scale graph-text pair pretraining provides a crucial initialization for learning generalizable and text-aligned graph representations. 
ii) \textbf{w/o rew} consistently underperforms the full model, suggesting that the proposed domain-aware reweighting helps the encoder better distinguish semantically distant inter-domain negatives while preserving intra-domain structure, thereby improving the quality of GNN representations for different domains.
iii) \textbf{w/o cur} also leads to noticeable degradation, implying that difficulty-aware curriculum reweighting during alignment is important for handling diverse datasets and improving overall alignment quality.

\subsection{Time and Resource Consumption}
\label{sec:time_and_resource}
Table~\ref{tab:efficiency} compares the time and memory consumption of \model and representative baselines under the cross-domain and cross-task generalization settings in Table~\ref{tab:cross}. To avoid affecting model performance, we retain the original batch size of each baseline whenever possible. GraphGPT, LLaGA, and GOFA are trained on the full Arxiv training set, whereas TEA-GLM and \model are trained on 40,000 graph instances sampled from the Arxiv training set. The results show that \model achieves competitive efficiency while delivering substantially stronger cross-domain and cross-task performance.
\begin{table}[!hbt]
\centering
\caption{Efficiency comparison of different methods across datasets, including runtime, batch size, and GPU memory usage.}
\resizebox{1\textwidth}{!}{
\begin{tabular}{lccccccccc}
\toprule

& \multicolumn{3}{c}{\textbf{Arxiv (Training)}} 
& \multicolumn{3}{c}{\textbf{Cora (Inference)}} 
& \multicolumn{3}{c}{\textbf{PubMed (Inference)}} \\

\cmidrule(lr){2-4} \cmidrule(lr){5-7} \cmidrule(lr){8-10}

Method
& Time (s) 
& Batch size 
& Mem (MB) 
& Time (s) 
& Batch size 
& Mem (MB) 
& Time (s) 
& Batch size 
& Mem (MB) 
\\

\midrule

GraphGPT
& 22641 & 1 & 30717
& 338 & 1 & 14550
& 8200 & 1 & 14651
\\

LLaGA
& 18286 & 16 & 34447 
& 3339 & 1 & 13222
& 19260 & 1 & 13215
\\

GOFA
& 137175 & 1 & 53100 
& 3385 & 1 &  22732
& 39961 & 1 &  23083
\\

TEA-GLM
& 4866 & 2 & 19941 
& 327 & 20 & 16142 
& 1075 & 20 & 16501 
\\

\midrule

Ours
& 11072 & 3 & 34953 
& 181 & 20 & 30891 
& 2136 & 20 & 32250 
\\

\bottomrule
\end{tabular}
}
\label{tab:efficiency}
\end{table}

\section{Conclusion}
In this paper, we study how to incorporate a multi-domain, multi-task GNN encoder into GLMs and align its representations with the LLM token space to produce unified graph tokens for diverse graph data. 
To this end, we propose \modelns, a unified graph language model that leverages multi-domain, multi-task graph-text pair pretraining to learn generalizable graph representations aligned with the textual modality, together with a curriculum alignment tuning mechanism that adaptively aligns GNN representations with LLMs according to per-domain alignment difficulty. Extensive experiments under multi-domain multi-task learning as well as cross-domain and cross-task generalization settings demonstrate that \model consistently outperforms state-of-the-art GLM baselines. A limitation of our current approach is that it focuses on conventional graph task types (node, edge, and graph classification) over a moderate collection of datasets, without explicitly handling more complex graph reasoning scenarios. Scaling the pretraining corpus to larger and more diverse graph-text collections and systematically studying how the resulting gains transfer to downstream tasks are important directions for future work.

\bibliographystyle{IEEEtran}
{
\small
\bibliography{main}
}

\newpage
\appendix

\section{Algorithm}
\label{app:algorithm}
We provide the complete training pipeline of \model in Algorithm~\ref{alg:alg1}.
\begin{algorithm}[!htb]
\centering
\caption{Training Pipeline of \modelns}
\label{alg:alg1}
\resizebox{\textwidth}{!}{%
\begin{minipage}{\textwidth}
\begin{algorithmic}[1]
\STATE \textbf{Input:} graph datasets $\mathcal{D}=\{D\}$; graph-text pair dataset $\mathcal{D}_{\mathrm{gt}}=\{(G_i,T_i,d_i)\}_{i=1}^{N}$; instruction-tuning dataset $\mathcal{D}_{\mathrm{it}}=\{(G_i,y_i,d_i)\}$; multi-scale GNN encoder $\mathrm{GNN}_{\phi}^{\mathrm{multi}}$; projector $\mathrm{Project}_{\theta}$; text encoder $\mathrm{Enc}_{\mathrm{text}}$; warmup ratio $\rho$; EMA momentum $\beta$; temperature $\tau$.
\STATE \textbf{Initialize:} GNN parameters $\phi$, projector parameters $\theta$, and domain difficulty statistics $\{\mu_D,\tilde g_D\}_{D\in\mathcal{D}}$.

\STATE \textit{Stage I: Graph-Text Pair Pretraining}
\STATE Compute domain centers and inter-domain distance matrices $M_g,M_t$ according to Eq.~\eqref{eq:domain_distance}.
\STATE Construct domain-aware weights $W^g,W^t$ according to Eq.~\eqref{eq:domain_weights}.
\FOR{each mini-batch $\mathcal{B}\subset\mathcal{D}_{\mathrm{gt}}$}
    \FOR{each graph-text pair $(G_i,T_i,d_i)\in\mathcal{B}$}
        \STATE Extract task-required graph representation $\mathbf{X}_i^{\ast}$ with $\mathrm{GNN}_{\phi}^{\mathrm{multi}}$ according to Eqs.~\eqref{eq:node_graph_representation}--\eqref{eq:graph-level}.
        \STATE Encode text representation $\mathbf{T}_i$ according to Eq.~\eqref{eq:text_representation}.
    \ENDFOR
    \STATE Compute bidirectional similarities $\mathbf{S}^{g\rightarrow t}$ and $\mathbf{S}^{t\rightarrow g}$ according to Eq.~\eqref{eq:similarity_matrices}.
    \STATE Compute the domain-reweighted contrastive loss $\mathcal{L}_{\mathrm{DR\mbox{-}CLIP}}$ according to Eq.~\eqref{eq:dr_clip_loss}.
    \STATE Update $\phi$ by minimizing $\mathcal{L}_{\mathrm{DR\mbox{-}CLIP}}$.
\ENDFOR

\STATE \textit{Stage II: Curriculum Alignment Tuning}
\STATE Set total training steps $T$ and warmup horizon $T_w \leftarrow \lfloor \rho T \rfloor$.
\FOR{optimization step $k=1,\dots,T$}
    \STATE Sample mini-batch $\mathcal{B}_k\subset\mathcal{D}_{\mathrm{it}}$ and identify active domains $\mathcal{D}_k$.
    \FOR{each sample $(G_i,y_i,d_i)\in\mathcal{B}_k$}
        \STATE Extract task-specific graph representation $\mathbf{X}_i^{\ast}$ using $\mathrm{GNN}_{\phi}^{\mathrm{multi}}$.
        \STATE Project $\mathbf{X}_i^{\ast}$ into graph tokens $\mathbf{z}_i^{\ast} \leftarrow \mathrm{Project}_{\theta}(\mathbf{X}_i^{\ast})$.
        \STATE Form graph-conditioned instruction $I_i=[\mathbf{q}_i;\mathbf{z}_i^{\ast}]$ and compute per-instance loss $L_i$ as in Eq.~\eqref{eq:instruction-tuning}.
    \ENDFOR
    \STATE Aggregate dataset-level losses $L_D^{(k)}$ according to Eq.~\eqref{eq:dataset_loss}.
    \STATE Estimate and smooth domain-level alignment difficulty using Eqs.~\eqref{eq:alignment_difficulty} and \eqref{eq:smoothed_difficulty}.
    \STATE Compute curriculum weights $w_D^{(k)}$ according to Eq.~\eqref{eq:curriculum_weights}.
    \STATE Compute curriculum-weighted objective $\mathcal{L}_k$ according to Eq.~\eqref{eq:curriculum_loss}.
    \STATE Update $\theta$ by minimizing $\mathcal{L}_k$.
\ENDFOR

\end{algorithmic}
\end{minipage}%
}
\end{algorithm}

\section{Instructions Details}

\subsection{Summarization Prompts}
\label{app:summarization_prompts}
We present the summarization prompts used to generate textual descriptions for graph instances from different graph datasets.
\begin{tcolorbox}[
    breakable,
    colback=gray!3,          
    colframe=myblue,      
    title=Summarization Prompts Details for Different Graph Datasets,
    coltitle=black,
    colbacktitle=lightblue,    
    fonttitle=\bfseries,
    boxrule=0.6pt,           
    arc=0pt,                 
    outer arc=0pt,          
    left=6pt,
    right=6pt,
    top=6pt,
    bottom=6pt
]

\textbf{Cora}: \\

I am providing you with a GraphML file depicting a citation network in artificial intelligence research. Each node in the network represents a scholarly article, and each edge signifies a citation relationship between articles. Please analyze the article represented by node `n\textcolor{purple}{\{node\}}' using the provided GraphML data in the following two ways:

\vspace{0.5em}
1. Paper Summary and Context Analysis:

- Extract and summarize the key findings or contributions of the paper denoted by `n\textcolor{purple}{\{node\}}'. Consider the details embedded within node `n\textcolor{purple}{\{node\}}', including its title, abstract, and keywords if available.

- Provide an overall summary of prevalent themes or concepts shared by the papers that cite or are cited by `n\textcolor{purple}{\{node\}}', namely its direct neighbors in the network. Identify common threads or research topics among these neighbors.

\vspace{0.5em}
2. Research Area Classification:

- Based on the information summarized from `n\textcolor{purple}{\{node\}}' and its neighboring nodes, determine the specific research area to which `n\textcolor{purple}{\{node\}}' primarily contributes.

- Justify the classification by explaining which aspects of `n\textcolor{purple}{\{node\}}' align with recognized themes, issues, or methodologies in the identified research area.

\vspace{0.5em}
Please ensure your analyses are grounded in the data provided by the GraphML file within 400 tokens, focusing on node `n\textcolor{purple}{\{node\}}' and its immediate citation neighborhood. The detailed GraphML citation network data is as follows:

\textcolor{purple}{\{GraphML\}}

\vspace{1em}

\textbf{Pubmed}: \\

I am providing you with a GraphML file depicting a citation network in medical research. Each node in the network represents a scholarly article, and each edge signifies a citation relationship between articles. Please analyze the article represented by node `n\textcolor{purple}{\{node\}}' using the provided GraphML data in the following two ways:

\vspace{0.5em}
1. Paper Summary and Context Analysis:

- Extract and summarize the key findings or contributions of the paper denoted by `n\textcolor{purple}{\{node\}}'. Consider the details embedded within node `n\textcolor{purple}{\{node\}}', including its title, abstract, and keywords if available.

- Provide an overall summary of prevalent themes or concepts shared by the papers that cite or are cited by `n\textcolor{purple}{\{node\}}', namely its direct neighbors in the network. Identify common threads or research topics among these neighbors.

\vspace{0.5em}
2. Research Area Classification:

- Based on the information summarized from `n\textcolor{purple}{\{node\}}' and its neighboring nodes, determine the specific research area to which `n\textcolor{purple}{\{node\}}' primarily contributes.

- Justify the classification by explaining which aspects of `n\textcolor{purple}{\{node\}}' align with recognized themes, issues, or methodologies in the identified research area.

\vspace{0.5em}
Please ensure your analyses are grounded in the data provided by the GraphML file within 400 tokens, focusing on node `n\textcolor{purple}{\{node\}}' and its immediate citation neighborhood. The detailed GraphML citation network data is as follows:

\textcolor{purple}{\{GraphML\}}

\vspace{1em}

\textbf{Arxiv}: \\

I am providing you with a GraphML file depicting a citation network in computer science research from arXiv. Each node in the network represents a scholarly article, and each edge signifies a citation relationship between articles. Please analyze the article represented by node `n\textcolor{purple}{\{node\}}' using the provided GraphML data in the following two ways:

\vspace{0.5em}
1. Paper Summary and Context Analysis:

- Extract and summarize the key findings or contributions of the paper denoted by `n\textcolor{purple}{\{node\}}'. Consider the details embedded within node `n\textcolor{purple}{\{node\}}', including its title, abstract, and keywords if available.

- Provide an overall summary of prevalent themes or concepts shared by the papers that cite or are cited by `n\textcolor{purple}{\{node\}}', namely its direct neighbors in the network. Identify common threads or research topics among these neighbors.

\vspace{0.5em}
2. Research Area Classification:

- Based on the information summarized from `n\textcolor{purple}{\{node\}}' and its neighboring nodes, determine the specific research area to which `n\textcolor{purple}{\{node\}}' primarily contributes.

- Justify the classification by explaining which aspects of `n\textcolor{purple}{\{node\}}' align with recognized themes, issues, or methodologies in the identified research area.

\vspace{0.5em}
Please ensure your analyses are grounded in the data provided by the GraphML file within 400 tokens, focusing on node `n\textcolor{purple}{\{node\}}' and its immediate citation neighborhood. The detailed GraphML citation network data is as follows:

\textcolor{purple}{\{GraphML\}}

\vspace{1em}

\textbf{WikiCS}: \\

I am providing you with a GraphML file that describes a hyperlink-based graph constructed from Wikipedia articles in the field of computer science. Each node represents a Wikipedia entry, and each edge represents a hyperlink relationship between entries.

\vspace{0.5em}
Please analyze the Wikipedia entry represented by node `n\textcolor{purple}{\{node\}}' based on the provided GraphML data, and complete the following tasks:

\vspace{0.5em}
1. Content Understanding and Neighborhood Analysis:

- Summarize the main topic and key technical concepts of the Wikipedia entry denoted by `n\textcolor{purple}{\{node\}}', using the entry name and entry content provided in the node.

- Then, examine the immediate neighboring nodes of `n\textcolor{purple}{\{node\}}' in the graph. Identify common themes, technologies, systems, or research topics that appear among these neighboring entries, and describe how they are conceptually related.

\vspace{0.5em}
2. Topic Classification:

- Based on the content of `n\textcolor{purple}{\{node\}}' and the semantic context provided by its neighboring entries, determine the single most appropriate category for `n\textcolor{purple}{\{node\}}'.

- Provide a brief justification explaining why the selected category best fits the topic and technical focus of `n\textcolor{purple}{\{node\}}'.

\vspace{0.5em}
Please ensure your analysis is grounded solely in the provided GraphML data, focuses on node `n\textcolor{purple}{\{node\}}' and its immediate neighbors, and is within 400 tokens.

The GraphML data is as follows:

\textcolor{purple}{\{GraphML\}}

\vspace{1em}

\textbf{WN18RR, FB15K237}: \\

I am providing you with a GraphML file that represents a local subgraph from a knowledge graph. Each node corresponds to an entity, and each directed edge represents a semantic relationship between entities.

\vspace{0.5em}
Please analyze the link specified by the entity pair (head, tail) corresponding to nodes `n\textcolor{purple}{\{head\}}' and `n\textcolor{purple}{\{tail\}}', using the provided GraphML data, and complete the following tasks:

\vspace{0.5em}
1. Link Semantic Analysis:

- Summarize the semantic meaning of the relationship between the head entity and the tail entity, based on their entity names, descriptions, and the surrounding graph structure. Consider how the head and tail entities are conceptually related.

\vspace{0.5em}
2. Relation Type Classification:

- Based on your analysis, determine the most appropriate relation type between the head and tail entities.

- Provide a brief justification explaining why this relation best describes the semantic connection between the two entities.

\vspace{0.5em}
Please ensure that your analysis is grounded solely in the provided GraphML data and focuses on the specified link and its local neighborhood, within 400 tokens.

The GraphML data is as follows:

\textcolor{purple}{\{GraphML\}}

\vspace{1em}

\textbf{ChemHIV, ChemPCBA}: \\

I am providing you with a GraphML file that represents a molecular graph. Each node corresponds to an atom with detailed atomic properties, and each edge represents a chemical bond or structural connection between atoms.

\vspace{0.5em}
Please analyze the molecular graph using the provided GraphML data and generate a summary by completing the following tasks:

\vspace{0.5em}
1. Structural Description and Characterization:

- Describe the molecule's overall structure, such as whether it is linear, cyclic, or branched, and its approximate size. Identify and describe the most important functional groups, special bonding patterns, and any notable structural features.

\vspace{0.5em}
2. Structural Insights for Classification:

- Based on the structural features identified above, infer the most likely chemical property or compound class of this molecule that could be relevant for graph-level classification tasks.

- Provide a brief explanation why these structural characteristics lead to your inference.

\vspace{0.5em}
Please ensure that your summary is grounded solely in the provided GraphML data and focuses on the overall structure and properties of the entire graph. The summary should be no more than 400 tokens.

The molecular GraphML data is as follows:

\textcolor{purple}{\{GraphML\}}

\vspace{1em}

\textbf{Case of \textcolor{purple}{\{GraphML\}}}: \\
\vspace{-1em}
\begin{verbatim}
<?xml version="1.0" encoding="UTF-8"?>
<graphml>
    <key id="k0" for="node" attr.name="title" attr.type="string">
        <default>unkown</default>
    </key>
    <key id="k1" for="node" attr.name="description" attr.type="string"/>
    <key id="k2" for="edge" attr.name="relation" attr.type="string"/>
    <graph id="G" edgedefault="undirected">
        <node id="n2">
            <data key="k0">M.L.(1996) Design by ...</data>
            <data key="k1">Designers rely on past experience ...</data>
        </node>
        <node id="n172">
            <data key="k0">Developing Case-Based Reasoning ...</data>
            <data key="k1">Case-based reasoning ...</data>
        </node>
        <edge id="e0" source="n172" target="n2">
            <data key="k2">cites</data>
        </edge>
        <edge id="e1" source="n2" target="n172">
            <data key="k2">cites</data>
        </edge>
    </graph>
</graphml>
\end{verbatim}

\vspace{1em}

\textbf{Case of Summary}: \\

1. Paper Summary and Context Analysis:

The paper represented by node n2 is titled ``M.L. (1996) Design by Interactive Exploration Using Memory-Based Techniques.'' It discusses the use of memory-based techniques from artificial intelligence to aid in the design process, particularly in the domain of structural design. The paper introduces DEMEX, an interactive computer-aided design system that leverages memory-based techniques to help users explore and refine design problems. The key contribution is the integration of memory-based methods to support exploration and understanding of ill-structured design problems.

The paper's neighbors include n172, titled ``Developing Case-Based Reasoning for Structural Design.'' This paper explores the development of case-based reasoning systems for design, including the implementation of four systems: CASECAD, CADSYN, WIN, and DEMEX. It emphasizes the use of case memory organization, recall, and adaptation in design tasks.

The shared themes between n2 and n172 include:
- Memory-based techniques for design.
- Case-based reasoning as a method for supporting design tasks.
- Interactive exploration in the early stages of design.
- Structural design as the application domain.
- Adaptation of design cases to fit new contexts.

These themes suggest a strong focus on case-based reasoning and interactive design exploration in the context of structural design.

2. Research Area Classification:

Based on the analysis, node n2 primarily contributes to the research area ``Case\_Based''.

Justification:
- The paper introduces DEMEX, a system that uses memory-based techniques to support design exploration, which aligns with case-based reasoning.
- The paper is cited by and cites n172, which explicitly discusses the development of case-based reasoning systems for structural design.
- The shared themes of case memory organization, case recall, and case adaptation are central to case-based reasoning.
- The focus on interactive exploration and design problem refinement is consistent with the broader goals of case-based reasoning in design.

Thus, n2 is best classified under ``Case\_Based'' due to its direct contribution to the development and application of case-based reasoning techniques in design.

\end{tcolorbox}

\subsection{Instruction Tuning Prompts}
\label{app:instruction_prompts}
We present the instruction tuning prompts used to align graph representations with LLMs across different graph datasets. Notably, for \textit{FB15K237}, including all 237 relation candidate labels in the instruction would result in excessively long input sequences for all GLMs. To reduce the token length, we adopt a 10-way sampling strategy: each instruction includes 10 candidate labels, consisting of the correct label and 9 negative labels randomly sampled from the remaining labels.

\begin{tcolorbox}[
    breakable,
    colback=gray!3,          
    colframe=myblue,      
    title=Instruction Tuning Prompts Details for Different Graph Datasets,
    coltitle=black,
    colbacktitle=lightblue,    
    fonttitle=\bfseries,
    boxrule=0.6pt,           
    arc=0pt,                 
    outer arc=0pt,          
    left=6pt,
    right=6pt,
    top=6pt,
    bottom=6pt
]

\textbf{Cora}: \\

Given a representation of a paper: <Node 1>, with the following information:

Title: \textcolor{purple}{\{title\}}

Abstract: \textcolor{purple}{\{abstract\}}

Question: Which machine learning sub-category does this paper belong to? Please directly give the most likely answer from the following sub-categories: \textcolor{purple}{\{candidate\_labels\}}.

\vspace{1em}

\textbf{PubMed}: \\

Given a representation of a paper: <Node 1>, with the following information:

Title: \textcolor{purple}{\{title\}}

Abstract: \textcolor{purple}{\{abstract\}}

Question: Which Medical Subject Headings (MeSH) term does this paper belong to? Please directly give the most likely answer from the following MeSH terms: \textcolor{purple}{\{candidate\_labels\}}.

\vspace{1em}

\textbf{Arxiv}: \\

Given a representation of a paper: <Node 1>, with the following information:

Title: \textcolor{purple}{\{title\}}

Abstract: \textcolor{purple}{\{abstract\}}

Question: Which arXiv CS sub-category does this paper belong to? Please directly give the most likely answer from the following sub-categories: \textcolor{purple}{\{candidate\_labels\}}.

\vspace{1em}

\textbf{WikiCS}: \\

Given a representation of a Wikipedia page: <Node 1>, with the following information:

Name: \textcolor{purple}{\{name\}}.

Content: \textcolor{purple}{\{content\}}.

Question: Which category does this Wikipedia page belong to? Please directly give the most likely answer from the following categories: \textcolor{purple}{\{candidate\_labels\}}.

\vspace{1em}

\textbf{WN18RR, FB15K237}: \\

Given a representation of relation between two entities: <Node 1>, with the following information:

First entity: Name: \textcolor{purple}{\{head\_name\}}, Description: \textcolor{purple}{\{head\_description\}}.

Second entity: Name: \textcolor{purple}{\{tail\_name\}}, Description: \textcolor{purple}{\{tail\_description\}}.

Question: Which category should the relation between these two entities be classified as? Please directly give the most likely answer from the following categories: \textcolor{purple}{\{candidate\_labels\}}.

\vspace{1em}

\textbf{ChemHIV, ChemPCBA}: \\

Given a representation of a molecule <Node 1>, with the following information:

SMILES: \textcolor{purple}{\{smiles\}}.

Question: \textcolor{purple}{\{task\}} Please answer: ``Yes, this molecule is effective to this assay'' or ``No, this molecule is not effective to this assay''.

\end{tcolorbox}

\section{Experiment Details and Additional Results}
\label{app:exp_details}

\subsection{Datasets Details}
\label{app:datasets}

\paragraph{Dataset Statistics.} We adopt the preprocessing pipeline of \cite{wang2024gft}, where raw textual descriptions associated with nodes and edges are encoded into 768-dimensional representations using a Sentence-BERT~\cite{reimers2019sentence}. For knowledge graphs (KGs), edge textual information is not further encoded as features, since the original textual content is already sufficient for KG completion tasks. Detailed dataset statistics are summarized in Table~\ref{tab:dataset_stats} and describe the dataset details as follows.
 
\begin{table}[!ht]
    \centering
      \caption{Dataset statistics. The number of graph instances depends on the task type: for node classification, it corresponds to the number of nodes; for edge classification, it corresponds to the number of edges; and for graph classification, it equals the number of graphs.}

    \label{tab:dataset_stats}
  
    \resizebox{1\textwidth}{!}{
      \begin{tabular}{lcccccc}
        \toprule
        \textbf{Dataset}  & \textbf{Task} & \textbf{\# Graphs} & \textbf{Avg. \# Nodes} & \textbf{Avg. \# Edges} & \textbf{\# Graph Instances} & \textbf{\# Classes} \\ \midrule
        \textbf{Cora}     & Node          & 1                  & 2,708                 & 10,556                & 2,708                      & 7                   \\
        \textbf{PubMed}   & Node          & 1                  & 19,717                & 44,338                & 19,717                     & 3                   \\
        \textbf{Arxiv}    & Node          & 1                  & 169,343               & 1,166,243             & 169,343                    & 40                  \\
        \textbf{Wiki-CS}  & Node          & 1                  & 11,701                & 216,123               & 11,701                     & 10                  \\
        \textbf{FB15K237} & Link          & 1                  & 14,541                & 310,116               & 310,116                    & 237                 \\
        \textbf{WN18RR}   & Link          & 1                  & 40,943                & 93,003                & 93,003                     & 11                  \\
        \textbf{ChemPCBA} & Graph         & 437,929            & 26.0                  & 28.1                  & 437,929                    & 128                 \\
        \textbf{ChemHIV}  & Graph         & 41,127             & 25.5                  & 27.5                  & 41,127                     & 2                   \\
        \bottomrule
      \end{tabular}
    }

\end{table}

\paragraph{Dataset Splitting.} 
\label{app:dataset_splitting}
We follow the same data splitting strategy as~\cite{liu2023one}. For datasets with multiple predefined splits, we adopt the first split for consistency. 
During the graph-text pair pretraining stage, we utilize all available data to train the GNN encoder. In the instruction tuning stage, we train on the training split of each dataset. For large-scale datasets such as \textit{Arxiv} and \textit{FB15K237}, we randomly sample subsets for training (Arxiv: 40,000; FB15K237: 30,000), while for smaller datasets, we use the full training set. 
Notably, the training sets of \textit{ChemPCBA} and \textit{ChemHIV} exhibit severe class imbalance, which can degrade the performance of all methods. Since handling class imbalance is not the primary focus of this work or GLM-based approaches, we apply balanced sampling subset from the training set to ensure the model can learn meaningful signals.
For datasets with a limited number of graph instances, we increase the training data through repeated sampling, using larger effective training sizes (Cora: 20$\times$, PubMed: 20$\times$, ChemHIV: 10$\times$). 
All methods use the aforementioned training data for instruction tuning. 
During evaluation, we report performance on the test split of each dataset. 

\paragraph{Evaluation Metrics.}
For node- and edge-level classification, we report classification accuracy (ACC\%) and F1-score (F1\%), for graph-level tasks on molecular datasets, we report the area under the ROC curve (AUC\%). 

\subsection{Baselines and Implementation Details}
\label{app:baselines}
We compare \model with a range of baselines, including pure LLMs and state-of-the-art GLMs. For a fair comparison, we use the same LLM backbone, Vicuna-7B-v1.5~\cite{vicuna2023}, for our model and all baselines except GOFA, which uses Mistral-7B-Instruct-v0.2~\cite{jiang2024mixtral}. Under the multi-domain, multi-task learning setting, for baselines that require retraining, we use pretraining data of the same scale and apply the same repeated sampling strategy as described in Appendix~\ref{app:dataset_splitting}. Under the cross-domain and cross-task generalization settings, GraphGPT, GOFA, and LLaGA are trained on the full Arxiv training set, whereas TEA-GLM and \model are trained on 40,000 graph instances sampled from the Arxiv training set.

\begin{itemize}[leftmargin=0.5cm]
\item \textbf{LLM Baselines.} We evaluate the performance of pure LLMs without any graph-specific encoding or alignment, including Vicuna-7B-v1.5~\cite{vicuna2023} and LLaMA-2-7B~\cite{touvron2023llama}. These models are directly prompted with natural language descriptions of the graph tasks without any additional graph representation. 
\item \textbf{GraphGPT~\cite{tang2024graphgpt}.}
GraphGPT is a graph-oriented instruction-tuned framework that augments a pre-trained LLM with graph structural representations. It aligns graph structural information with the natural language space through text-graph grounding, and maps graph embeddings into LLM-compatible graph tokens via a lightweight alignment projector within a two-stage graph instruction tuning paradigm. Since its released training and evaluation pipelines are primarily designed for node-level presentations, we pretrain GraphGPT on all node-level datasets and restrict its evaluation to these datasets.
\item \textbf{LLaGA~\cite{chen2024llaga}. } 
LLaGA adapts graph data into structure-aware node sequences that preserve structural information, which are then projected into the LLM token embedding space via a projector. Since the original implementation of LLaGA only supports node-level presentations, we pretrain it on all node-level datasets and restrict evaluation to these datasets.
\item \textbf{GOFA~\cite{konggofa}.} 
GOFA proposes a generative graph language model by interleaving GNN layers with a pre-trained LLM, enabling joint modeling of graph structure and textual semantics. It is pre-trained with a unified generative objective over multiple graph-related tasks, such as graph-level prediction and question answering. Due to the substantial computational cost of pretraining GOFA from scratch, we directly adopt the released checkpoints from the original paper for evaluation under the multi-domain, multi-task learning setting and convert the datasets used in our paper into the input format required by GOFA.

\item \textbf{TEA-GLM~\cite{wang2024llms}.} 
TEA-GLM introduces a text-enhanced graph alignment framework that aligns GNN-encoded graph representations with LLM token embeddings via instruction tuning, leveraging textual information to bridge graph and language modalities. The original implementation only supports node-level presentations. We extend it in a principled manner to support edge-level and graph-level tasks, and then retrain its GNN encoder and projector on multi-task, multi-domain data so that evaluation can be carried out across all task types.
\end{itemize}

\subsection{Complete Main Results}
We present the complete main results as in Table~\ref{tab:complete_main_results}, including accuracy and F1-score for node- and edge-level tasks, as well as AUC for graph-level tasks.
\begin{table}[!htbp]
\centering
\caption{Complete performance comparison of different methods on node classification, edge classification, and graph classification tasks under multi-domain and multi-task learning. The highest result is \textbf{bold}, and the second highest result is \underline{underlined}.}
\resizebox{\textwidth}{!}{
\begin{tabular}{lcccccccccccccc}
\toprule

& \multicolumn{8}{c}{\textbf{Node Classification}} 
& \multicolumn{4}{c}{\textbf{Edge Classification}} 
& \multicolumn{2}{c}{\textbf{Graph Classification}} \\

\cmidrule(lr){2-9} \cmidrule(lr){10-13} \cmidrule(lr){14-15}

Method
& \multicolumn{2}{c}{Cora}
& \multicolumn{2}{c}{PubMed}
& \multicolumn{2}{c}{Wiki-CS}
& \multicolumn{2}{c}{Arxiv}
& \multicolumn{2}{c}{WN18RR}
& \multicolumn{2}{c}{FB15K237}
& ChemPCBA
& ChemHIV
\\

\cmidrule(lr){2-3} \cmidrule(lr){4-5} \cmidrule(lr){6-7} \cmidrule(lr){8-9}
\cmidrule(lr){10-11} \cmidrule(lr){12-13} \cmidrule(lr){14-14} \cmidrule(lr){15-15}

& Acc & F1
& Acc & F1
& Acc & F1
& Acc & F1
& Acc & F1
& Acc & F1
& AUC
& AUC
\\

\midrule

Vicuna-7B
& 56.67 & 58.52
& 78.92 & 78.85
& 62.34 & 60.57
& 43.75 & 23.08
& 34.84 & 12.14
& 56.01 & 38.46
& 46.47
& 30.36
\\

LLaMA-2-7B
& 29.45 & 30.13
& 25.36 & 19.66
& 22.22 & 21.46
& 6.33 & 3.40
& 29.16 & 7.54
& 20.60 & 11.30
& 51.12
& 50.09
\\

GraphGPT
& 57.59 & 57.91
& 73.45 & 72.85
& 49.20 & 36.87
& 67.63 & 34.73
& N.S. & N.S.
& N.S. & N.S.
& N.S.
& N.S.
\\

LLaGA
& \underline{74.71} & \underline{73.29}
& 68.54 & 67.96
& \underline{69.16} & \underline{63.08}
& \underline{73.83} & \underline{55.86}
& N.S. & N.S.
& N.S. & N.S.
& N.S.
& N.S.
\\

GOFA
& 63.88 & 60.88
& 64.72 & 48.46
& 66.77 & 62.85
& 66.11 & 46.16
& 21.95 & 10.00
& 54.41 & 33.94
& 60.61
& 49.78
\\

TEA-GLM
& 62.38 & 59.71
& \underline{85.13} & \underline{84.32}
& 55.19 & 50.27
& 66.81 & 44.03
& \underline{92.85} & \underline{73.82}
& \underline{91.85} & \textbf{75.21}
& \underline{80.40}
& \underline{67.13}
\\

\midrule

Ours
& \textbf{76.98} & \textbf{75.72}
& \textbf{90.49} & \textbf{90.09}
& \textbf{76.65} & \textbf{72.40}
& \textbf{74.22} & \textbf{56.03}
& \textbf{93.52} & \textbf{75.30}
& \textbf{91.96} & \underline{74.00}
& \textbf{81.69}
& \textbf{70.09}
\\

\bottomrule
\end{tabular}
}
\label{tab:complete_main_results}
\end{table}

\subsection{Hyperparameters Analysis}
\label{app:hyperparameters}

\paragraph{Graph Token Length.}
\begin{figure*}[!htbp]
    \centering
    \subfloat[PubMed]{\includegraphics[width=0.33\textwidth]{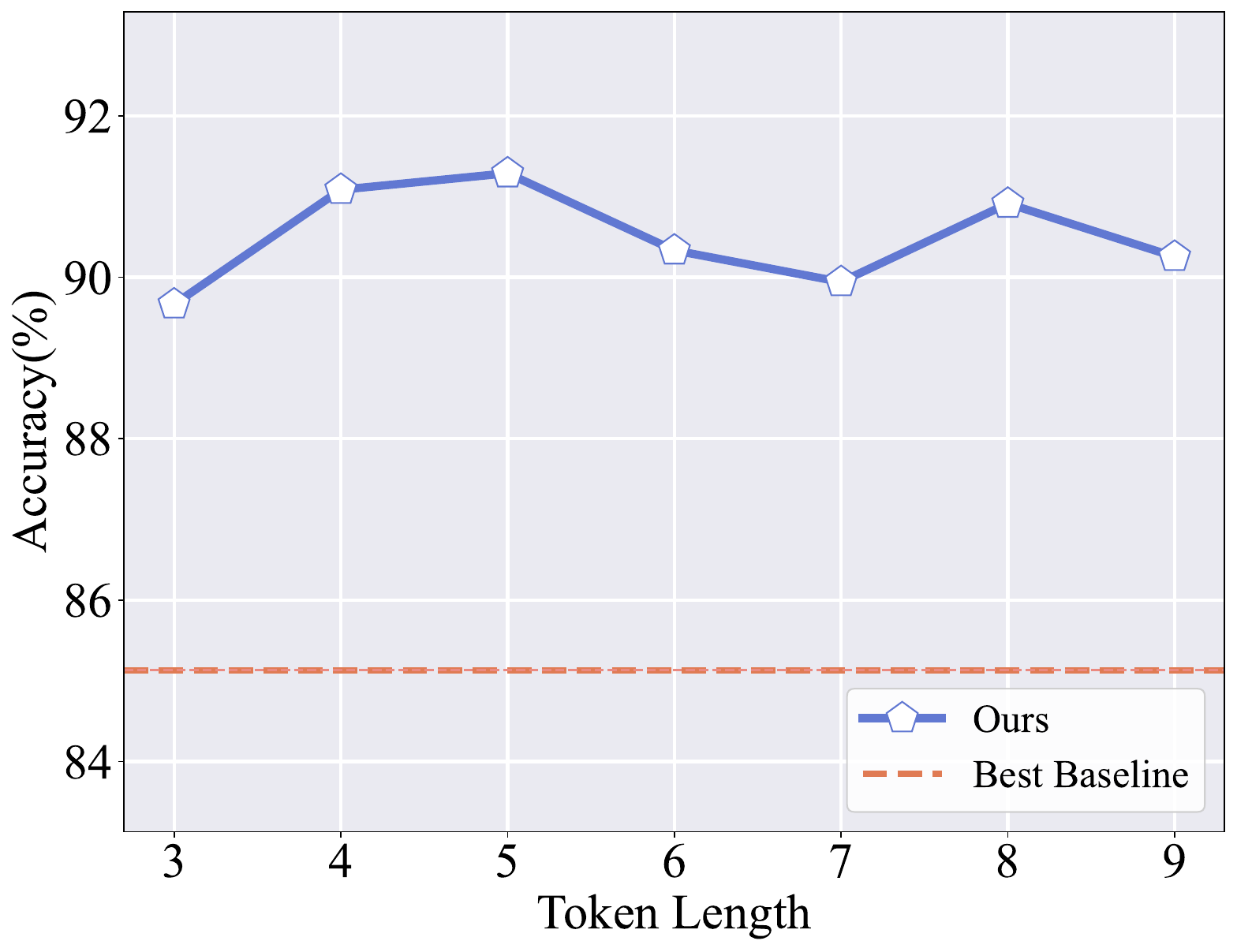}}
    \subfloat[Wiki-CS]{\includegraphics[width=0.33\textwidth]{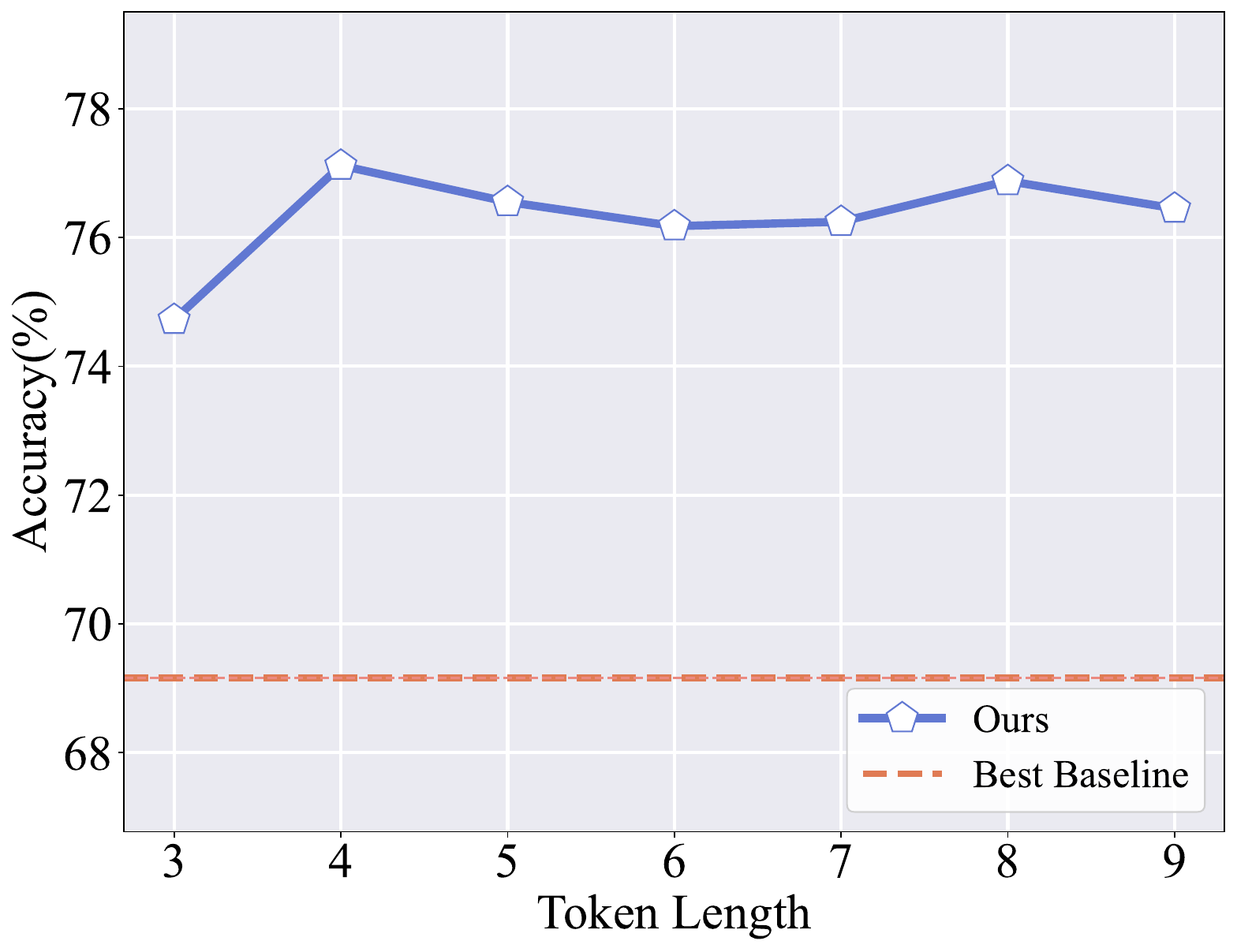}}
    \subfloat[ChemHIV]{\includegraphics[width=0.33\textwidth]{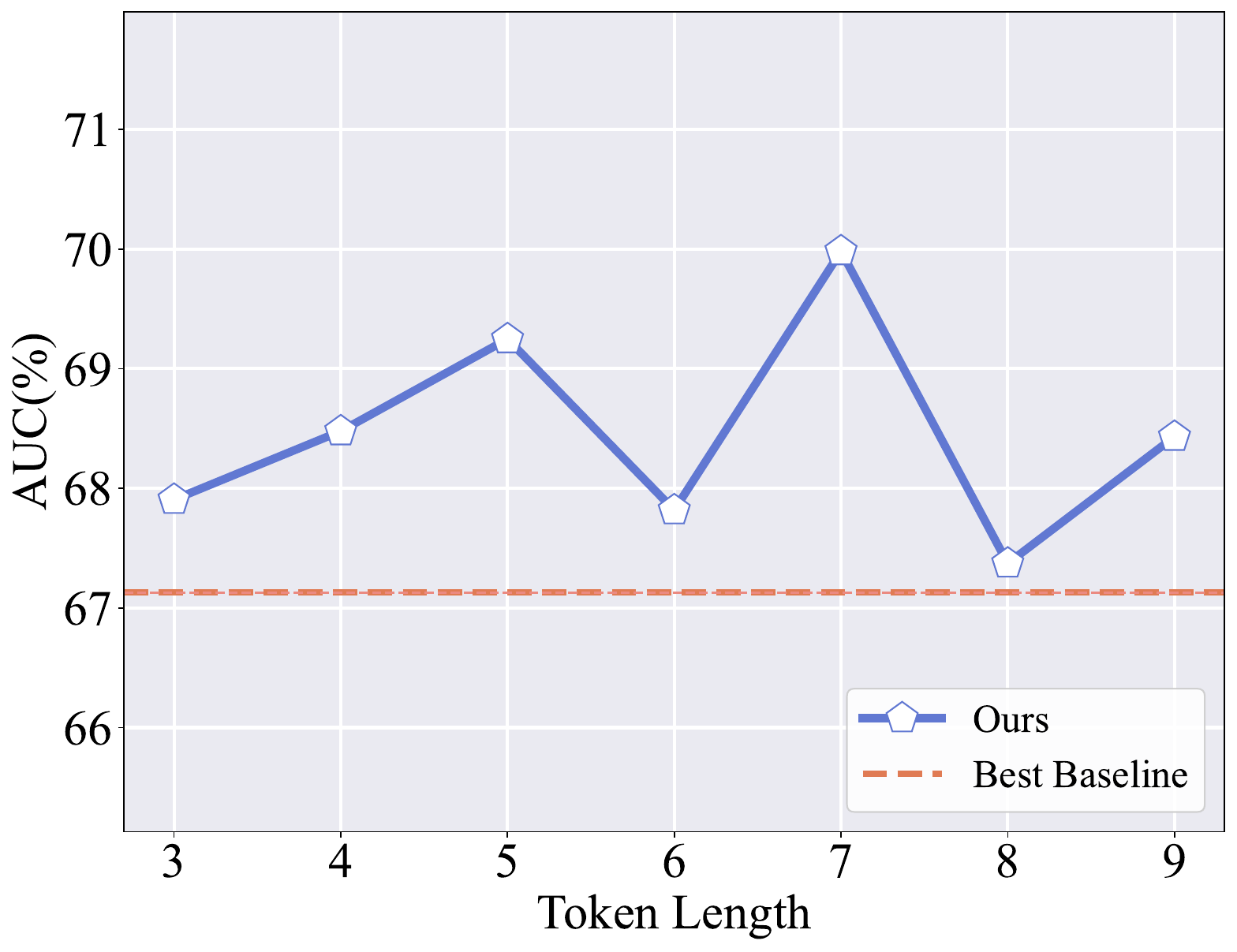}}
    \caption{Hyperparameter analysis of the graph token length $m$. The solid line denotes the performance of \modelns, while the dashed line indicates the best-performing baseline.}
    \label{fig:hyper_token_num}
\end{figure*}
We study the sensitivity of \model to the number of graph tokens $m$ injected into the LLM. Intuitively, a larger $m$ provides a higher-capacity interface for conveying fine-grained graph information (e.g., more queried nodes/edges or richer multi-scale features) to the LLM, but also increases the compute and memory cost due to the quadratic attention complexity in the input length (Sec.~\ref{app:time_complexity}). We vary $m$ in a moderate range (e.g., $m\in\{3,4,5,6,7,8,9\}$) while keeping other settings fixed, and report the resulting accuracy or AUC on several datasets. As shown in Figure~\ref{fig:hyper_token_num}, performance generally improves as $m$ grows from very small values and then saturates, with overly large $m$ yielding diminishing performance. Across the tested range, \model consistently outperforms the best baseline, indicating that its advantage is robust to the choice of graph token length.

\paragraph{Curriculum EMA Momentum.}
\begin{figure*}[!htbp]
    \centering
    \subfloat[PubMed]{\includegraphics[width=0.33\textwidth]{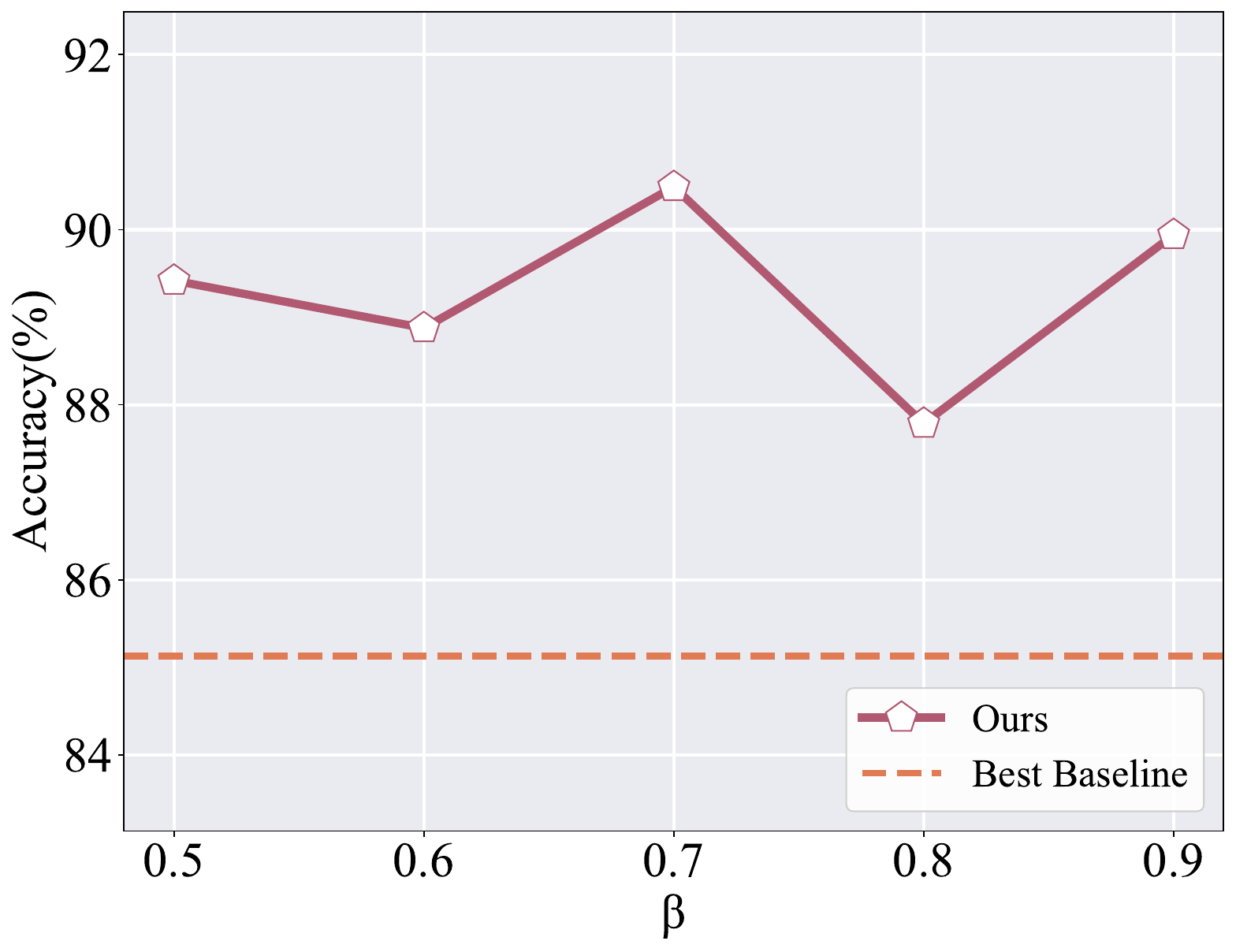}}
    \subfloat[Wiki-CS]{\includegraphics[width=0.33\textwidth]{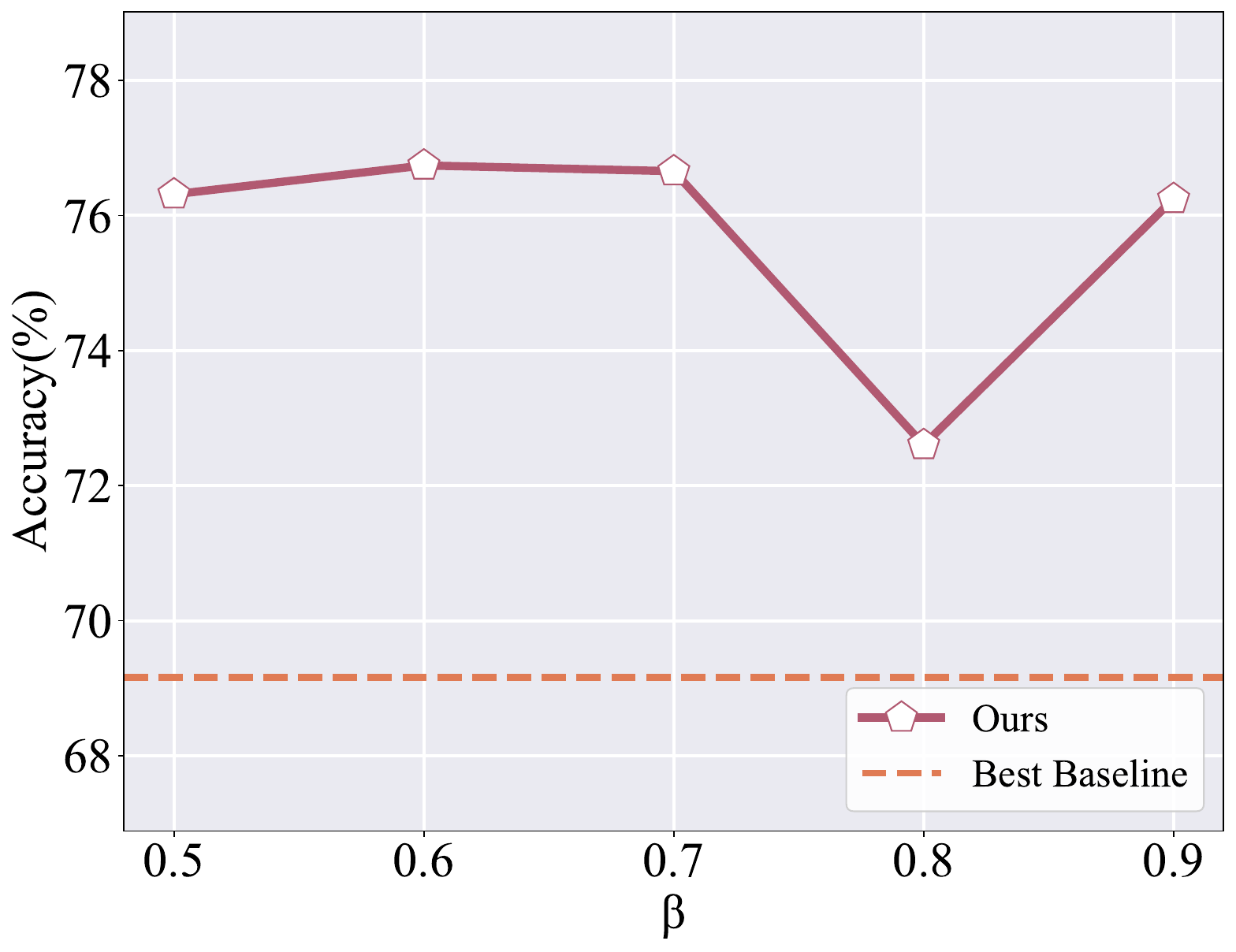}}
    \subfloat[ChemHIV]{\includegraphics[width=0.33\textwidth]{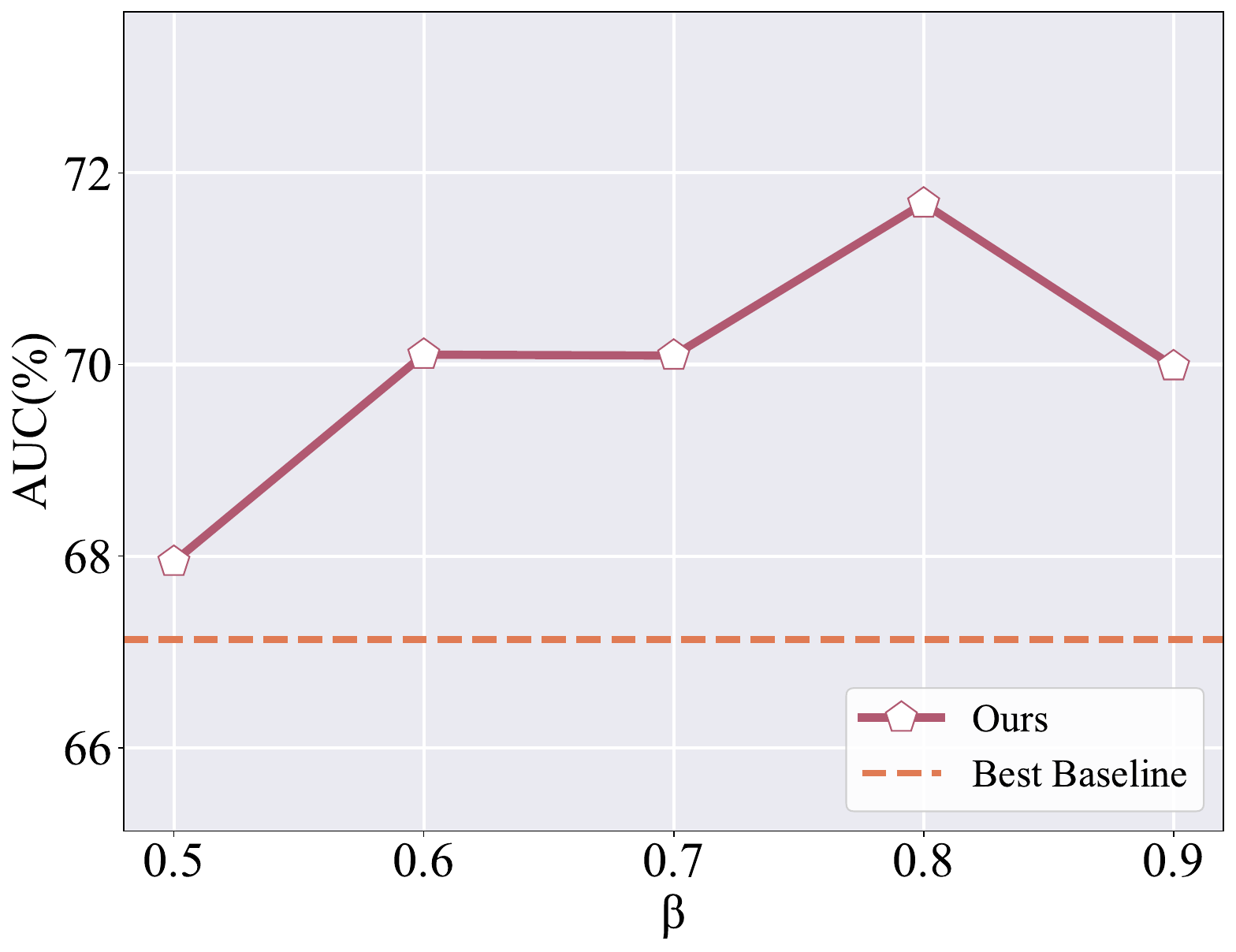}}
    \caption{Hyperparameter analysis of the EMA momentum $\beta$. The solid line denotes the performance of \modelns, while the dashed line indicates the best-performing baseline.}
    \label{fig:hyper_ema}
\end{figure*}
We further analyze the sensitivity of the curriculum alignment tuning to the EMA momentum $\beta$ in Eq.~\eqref{eq:smoothed_difficulty}, which controls the smoothness of the domain difficulty estimate $\tilde g_D^{(k)}$. A small $\beta$ makes the estimate highly responsive but potentially noisy, leading to unstable domain reweighting; a large $\beta$ yields a smoother curriculum but may react slowly to shifts in optimization dynamics, especially during early training. We vary $\beta$ over a broad range (e.g., $\beta\in\{0.5,0.6,0.7,0.8,0.9\}$) and report the final performance in Figure~\ref{fig:hyper_ema}. \model remains stable across a wide range of $\beta$ and consistently outperforms the best baseline, suggesting that the curriculum mechanism is not overly sensitive to this hyperparameter.

\subsection{Time Complexity Analysis}
\label{app:time_complexity}
We analyze the per-instance time complexity for a single graph-conditioned input $(G, I)$, where $G=(\mathcal{V},\mathcal{E},\mathbf{X},\mathbf{E})$ denotes the input graph and $I$ is the corresponding instruction with text length $L_t$. For a message-passing GNN with $L_g$ layers and hidden dimension $d_g$, each layer incurs $\mathcal{O}(|\mathcal{E}|d_g)$ for neighborhood aggregation and $\mathcal{O}(|\mathcal{V}|d_g^2)$ for node-wise transformations, resulting in an overall complexity of $\mathcal{O}\big(L_g(|\mathcal{E}|d_g + |\mathcal{V}|d_g^2)\big)$ for computing node representations. A global pooling operation further adds $\mathcal{O}(|\mathcal{V}|d_g)$ to obtain a graph-level representation. Task-specific representations (Eqs.~\ref{eq:node-level}--\ref{eq:graph-level}) are then constructed via a $2d_g \rightarrow d_g$ linear transformation over concatenated graph features, incurring a cost of $\mathcal{O}(d_g^2)$. These representations are subsequently projected into $m$ graph tokens in the LLM embedding space of dimension $d_l$, which requires $\mathcal{O}(m d_g d_l)$. Finally, given a text sequence of length $L_t$ augmented with $m$ graph tokens, a Transformer LLM with $L_l$ layers has a dominant self-attention complexity of $\mathcal{O}\big(L_l (L_t + m)^2 d_l\big)$. Overall, the end-to-end time complexity is $\mathcal{O}\big(L_g(|\mathcal{E}|d_g + |\mathcal{V}|d_g^2) + m d_g d_l + L_l (L_t + m)^2 d_l\big)$.

\subsection{Implementation Details}
\label{app:hyperparameters_details}
During graph-text pair pretraining, we use GraphSAGE~\cite{hamilton2017inductive} as the GNN backbone, with 3 layers and 768 for the input, hidden, and output dimensions. The multi-scale GNN encoder is trained for 100 epochs with a batch size of 4096 and a learning rate of $1\times10^{-4}$. For instruction tuning, we initialize the GNN encoder from the pretrained checkpoint and use Vicuna-7B-v1.5~\cite{vicuna2023} as the LLM backbone. The LLM backbone is frozen, and the alignment module is trained for one epoch with a batch size of 3 and a learning rate of 0.004. We set the graph token length $m$ to 7. For curriculum alignment tuning, we set the warmup ratio $\rho$ to 0.01 and the EMA momentum $\beta$ to 0.7. 
We pretrain the GNN encoder on 6 NVIDIA GeForce RTX 3090 GPUs, each with 24GB of memory, and perform instruction tuning on a single NVIDIA A100-SXM4-80GB GPU.

\section{More Related Works}
\label{app:related_works}
\paragraph{Multi-domain Multi-task Graph Learning.}
Recent mainstream efforts in multi-domain and multi-task graph learning have increasingly moved toward Graph Foundation Models (GFMs)~\cite{liu2023towards,sun2023all,he2025unigraph,sun2025riemanngfm,wang2025graph,wang2025towards,liu2025graph,yuan2026rag}, which aim to develop graph models that can generalize beyond a single dataset, domain, or task setting~\cite{yan2024inductive,yu2024hgprompt, yu2024multigprompt,shen2024zero,fang2025uniglm,zhao2025survey,yuan2025graver}. Instead of training separate models for each graph domain or downstream task, GFMs typically seek to share model parameters, representation spaces, prompting mechanisms, or pretraining objectives across heterogeneous graph data, thereby improving generalization and transferability~\cite{liu2023graphprompt,jiang2024ragraph,yu2025samgpt,yu2025non,zhu2025towards,yuan2026retrieving,yuan2025much}. Representative works have explored how to unify graph inputs, task formats, and transferable structural patterns across domains. For example, OFA~\cite{liu2023one} formulates diverse graph datasets as text-attributed graphs, where nodes and edges are described using natural language and encoded into a shared feature space. It further introduces nodes-of-interest and graph prompting mechanisms to standardize different graph classification tasks under a unified model. GFT~\cite{wang2024gft}, on the other hand, studies transferable graph patterns from the perspective of message passing by treating computation trees as reusable vocabulary tokens, enabling cross-domain and cross-task transfer through a shared tree vocabulary. Nevertheless, most existing methods focus on improving the generalization ability of graph encoders themselves, without explicitly studying how multi-domain, multi-task GNN representations should be aligned with LLMs to construct a graph language model. In contrast, our work targets this alignment problem by learning a shared GNN encoder from diverse graph-text pairs and adaptively aligning its representations with LLMs across domains and tasks.

\paragraph{Graph-Text Alignment.}
Graph-text alignment has been widely explored as an effective way to bridge graph-structured data and natural language semantics~\cite{ke2021jointgt, zhaolearning,chen2025curriculum}. A common strategy is to construct paired graph-text views, such as node-text, graph-text, or substructure-text pairs, and jointly train graph and text encoders with contrastive or matching objectives, so that structurally meaningful graph representations can be aligned with textual semantics~\cite{yang2021graphformers,zhang2025can,zhu2025llm}. Such alignment is particularly useful for text-attributed graphs and graph-language applications, where textual attributes provide rich semantic supervision beyond graph topology and task labels.
Representative methods mainly study graph-text alignment from the perspective of self-supervised representation learning. For example, ConGraT~\cite{brannon2024congrat} proposes a general contrastive graph-text pretraining framework for text-attributed graphs, where a language model and a graph neural network are trained to align text and node representations in a shared latent space through a CLIP-style contrastive objective. GraphCLIP~\cite{zhu2025graphclip} enhances the transferability of graph foundation models for text-attributed graphs by constructing LLM-generated graph-summary pairs and aligning graph and summary representations through self-supervised contrastive pretraining with invariant learning. 
However, existing graph-text alignment methods are typically designed for specific domains and tasks, and do not explicitly consider how to learn generalizable graph representations that can be aligned with LLMs across diverse domains and tasks. Our approach differs by using graph-text pair pretraining to obtain generalizable, text-aligned GNN representations and further performing curriculum alignment tuning to connect these representations with the LLM token space.


\end{document}